\documentclass[11pt,
  a4paper,
  parskip=half, 
  BCOR=10mm, 
  ngerman,
  english,
  oneside]{scrbook}
  \usepackage{tabularx}
\usepackage[a4paper,margin=1.3in,footskip=0.25in]{geometry}
\usepackage[ngerman, english]{babel} 
\usepackage[backend=bibtex]{biblatex}
\usepackage{titling}
\title{FinBERT-QA: Financial Question Answering with pre-trained BERT Language Models}
\author{Bithiah Yuan}
\date{July 31, 2020}  

\newcommand{\firstexaminer}{Dr. Fang Wei-Kleiner}
\newcommand{\secondexaminer}{Prof. Dr. Josif Grabocka}
\newcommand{\advisers}{Dr. Fang Wei-Kleiner}

    \usepackage[T1]{fontenc}  
    \usepackage[utf8]{inputenc}
    \usepackage{scrhack}

    \usepackage[labelfont=bf, labelsep=colon, format=hang, textfont=singlespacing]{caption}
    \usepackage{chngcntr}  
    \counterwithout{equation}{chapter}
    \counterwithout{figure}{chapter}
    \counterwithout{table}{chapter}

    \setkomafont{chapter}{\normalfont\bfseries\huge}

    \usepackage{setspace}  
    \usepackage[dvipsnames]{xcolor}  
    \usepackage{array}  
    \usepackage{graphicx}  
    \usepackage{subfig}  
    \usepackage{amsmath}  
    \usepackage{amsthm}   
    \usepackage{amsfonts} 
    \usepackage{calc}  
    \usepackage[unicode=true,bookmarks=true,bookmarksnumbered=true, bookmarksopen=true,bookmarksopenlevel=1,breaklinks=false, pdfborder={0 0 0},backref=false,colorlinks=false]{hyperref}
    \usepackage{etoolbox} 

    \usepackage{algorithm,algpseudocode}
    \usepackage{bm}  
    \usepackage{tikz}
    \usetikzlibrary{positioning}  
    
    \usepackage{ifthen}

    \usepackage[protrusion=true,expansion=true, kerning]{microtype}
    \usepackage{enumitem}
    
    \usepackage{booktabs}

    \usepackage{lipsum}

    \renewcommand{\eqref}[1]{Equation~(\ref{#1})}

    \definecolor{darkgreen}{rgb}{0.0, 0.5, 0.0}
    \definecolor{UniBlue}{RGB}{0, 74, 153}
    \definecolor{UniRed}{RGB}{193, 0, 42}
    \definecolor{UniGrey}{RGB}{154, 155, 156}

    \let\mySection\section\renewcommand{\section}{\suppressfloats[t]\mySection}
    
    \let\mySubSection\subsection\renewcommand{\subsection}{\suppressfloats[t]\mySubSection}

    \let\mySubsubSection\subsubsection\renewcommand{\subsubsection}{\suppressfloats[t]\mySubsubSection}

    \newcommand{\eqname}[1]{\tag*{#1}}

    \hypersetup{pdftitle={\thetitle}, hypertexnames=false}

    \tikzset{>=latex}  

    \algnewcommand\algorithmicforeach{\textbf{for}}
    \algdef{S}[FOR]{ForEach}[1]{\algorithmicforeach\ #1\ \algorithmicdo}

    \setlist{itemsep=.5em}
    
	\usepackage{ifthen} 
	\newcounter{todos}
	\newcounter{extends}
	\newcounter{drafts}
\begin{document}
    \pagestyle{empty} 
    \hypersetup{pageanchor=false}

    
\begin{titlepage}
\begin{center}

\newcommand{\HorizontalLine}{\rule{\linewidth}{0.3mm}}

{\Large Master's Thesis}\\[1.3cm]

\HorizontalLine \\[0.4cm]
{ \huge \bfseries \thetitle }
\HorizontalLine \\[1.5cm]

{\Huge \theauthor} \\[2cm]

\begin{tabular}[hc]{>{\huge}l >{\huge}l}
  Examiner: & \firstexaminer \\[0.3cm]
  Adviser: & \advisers \\[1.2cm]
\end{tabular}
\vfill  

\Large {
    Albert-Ludwigs-University Freiburg\\
    Faculty of Engineering\\
    Department of Computer Science\\
    Chair of Databases and Information Systems\\[1cm]

    July 31, 2020
 
}
\end{center}
\end{titlepage}

\thispagestyle{empty}
\ \vfill \ \\  
\
\textbf{Writing Period}            \smallskip{} \\
31.\,01.\,2020 -- 31.\,07.\,2020   \bigskip{} \\
\
\textbf{Examiner}                  \smallskip{} \\
\firstexaminer                     \bigskip{} \\
\
\ifdef{\secondexaminer}
	{
	\textbf{Second Examiner}       \smallskip{} \\
	\secondexaminer                \bigskip{} \\
	\
	}
	{
	}
\textbf{Adviser}                  \smallskip{} \\
\advisers

    \pagestyle{plain} 
    \frontmatter  

\chapter*{Declaration}

I hereby declare, that I am the sole author and composer of my thesis and that no other sources or learning aids, other than those listed, have been used. Furthermore, I declare that I have acknowledged the work of others by providing detailed references of said work.  \newline
I also hereby declare that my thesis has not been prepared for another examination
or assignment, either in its entirety or excerpts thereof. 
\\[3\normalbaselineskip]
\begin{tabular}{p{\textwidth/2} l}
  \rule{\textwidth/3}{0.4pt}   &   \rule{\textwidth/3}{0.4pt} \\
  Place, Date                  &   Signature
\end{tabular}

    \chapter*{Abstract}
Motivated by the emerging demand in the financial industry for the automatic analysis of unstructured and structured data at scale, Question Answering (QA) systems can provide lucrative and competitive advantages to companies by facilitating the decision making of financial advisers. Consequently, we propose a novel financial QA system using the Transformer-based pre-trained BERT language model to solve the limitations of data scarcity and language specificity in the financial domain. Our system focuses on financial non-factoid answer selection, which retrieves a set of passage-level texts and selects the most relevant as the answer. To increase the efficiency, we formulate the answer selection task as a re-ranking problem, in which our system consists of an \textit{Answer Retriever} using BM25, a simple information retrieval approach, to first return a list of candidate answers and an \textit{Answer Re-ranker} built with variants of pre-trained BERT language models to re-rank and select the most relevant answers. We investigate various learning, further pre-training, and fine-tuning approaches for BERT and our experiments suggest that FinBERT-QA, a model built from applying the Transfer and Adapt further fine-tuning and pointwise learning approach is the most effective as it improves the state-of-the-art results of task 2 of the FiQA dataset by 16\% on MRR, 17\% on NDCG, and 21\% on Precision@1.

    \chapter*{Acknowledgments}

I would like to thank:

\begin{itemize}
\item Dr. Fang Wei-Kleiner for providing me the freedom and encouragement to pursue a topic of my interest for this thesis. I am sincerely grateful for her feedback and  motivating advice. 
\item Prof. Dr. Josif Grabocka for agreeing to be my second examiner.
\item Dogu Araci, the creator of FinBERT, for the inspiration of my thesis as well as for helping me understand the theoretical concepts of BERT.
\item My proofreader: Robin Szekely
\item My family and friends for always being there for me and encouraging me throughout my thesis and the course of my studies.
\end{itemize}

\tableofcontents
    
\chapter{Introduction}\label{chap:Introduction}
\pagenumbering{arabic}

In recent years, Natural Language Processing (NLP) techniques have proven to be powerful due to their ability to automatically analyze domain-specific data at scale. The financial domain, in particular, relies on translating multiple unstructured and structured data for fast and detailed decision making \cite{boyer}. The use of sentiment analysis in the financial domain to predict the behavior of stock markets or credit risks have been widely researched and shown to be effective. This motivates further research of applying NLP methods such as Question Answering (QA) to the financial domain.

QA systems can be used to assist financial advisers of large institutions answer financial outlook and information questions from their clients. In order to answer questions that examine how financial entities, such as company stocks, currencies, industries, and assets are expected to perform in the future, the advisers need to provide textual passages with answers based on past or recent financial reports. However, retrieving financial reports pose a challenge since the experience levels of financial advisers vary, which results in inconsistent interpretations of the reports. Moreover, the large magnitude of detailed and technically written incoming financial reports are infeasible to read through even for the best and most experienced advisers \cite{boyer}.

As a result, the application of QA in the financial domain is critical due to the highly competitive and profitable nature of the industry. Since QA research targeting financial datasets is underexplored and large profits and competitive advantages can be achieved with slight improvements in the underlying technology \cite{desola}, further experimentation using NLP and Deep Learning approaches can greatly benefit financial industries in the current era of digital transformation. Following the need for a novel financial QA system, Section \ref{section:motivation} further reinforces our motivation, Section \ref{section:task} provides a formal task definition, and Section \ref{section:intro_approach} provides an overview of our approach and contributions.

\section{Motivation}\label{section:motivation}

Financial QA is more of a challenge compared to general QA due to the specialized terminology involved. For example, a sentence such as "there is an increase in the short sells of a company" requires domain-specific knowledge to know that it implies a negative outlook \cite{boyer}. Early NLP approaches for financial QA are based on feature engineering financial lexicons, for example, using handcrafted features based on existing financial knowledge to build a classifier predicting if an answer is relevant to a question \cite{boyer}. However, feature engineering and simple word counting are not enough since the correct answer may only be semantically related to the question. Moreover, the questions and answers may not share direct lexical units. For example, the correct answer could be formulated differently or serve to complete the question with missing information \cite{tran}. The correct answers can also be noisy and contain large amounts of irrelevant information and words that are not in the question \cite{qa-lstm}. As depicted in Table \ref{tab:QA}, which is an example of a question, ground truth answer, and an incorrect negative answer from the FiQA dataset \cite{fiqa}, QA in the financial domain is especially difficult because even though the negative answer has shared lexical units with the question such as "income" and "accrual", it does not answer the question semantically.

Deep learning models using neural networks have exploded in recent years and have shown to be effective as they require  little feature engineering and are able to better learn the context in a language corpus. However, traditional deep learning methods based on representation learning have the following drawbacks when applied to financial texts:

\begin{enumerate}
\item \textbf{Data Scarcity:} Deep learning models require a large amount of training data to understand the underlying latent patterns of the corpora. This presents a problem because large labeled financial datasets are scarce and building such a large-scale dataset would be complex and expensive since it requires annotations from highly trained financial experts \cite{araci}.
\item \textbf{Financial Language Specificity:} Financial corpora have specific language and underlying conceptualizations. Therefore, only training a model on the target corpus can fail to capture particular semantic and contextual meanings \cite{araci}.
\item \textbf{Shallow pre-trained Word Embeddings Usage:} Since many deep learning approaches are based on shallow pre-trained word embedding such as word2vec \cite{mikolov} or GloVe \cite{glove}, they are not able to deeply capture the contextual meanings of financial datasets \cite{araci}. 
\end{enumerate} 

Recently, there have been large developments in using transfer learning methods for NLP tasks in order to solve the above challenges. In particular, pre-trained language models such as ELMo \cite{elmo}, OpenAI GPT \cite{gpt} and BERT \cite{bert} have achieved several breakthroughs in the last two years. Pre-trained language models are extremely effective since they are first pre-trained on a large-scale corpora in an unsupervised manner, then the entire pre-trained model is applied to a downstream task to better capture the context and dependencies between words. The Transformer-based pre-trained language model, BERT, in particular, can be easily applied to different NLP tasks even if the size of the target task dataset is small to achieve the state-of-the-art results \cite{bert}. 

Since to our best knowledge no previous research using pre-trained language models has been conducted to address QA in the financial domain, the research interest of this thesis is to create novel systems using pre-trained BERT models that can capture the semantic relations of financial QA. We will formally define the task in the next section.

\begin{center}
\begin{minipage}{\linewidth}
\centering
\renewcommand{\arraystretch}{1.5}
\begin{tabular}{||>{\arraybackslash}p{12cm}||}

            \hline
            \textbf{Question:}  Are individual allowed to use accrual based accounting for federal income tax?\\ \hline
             \textbf{Ground truth answer:} Yes. But once you chose the method (on your first tax return), you cannot change it without the IRS approval. Similarly the fiscal year. For individuals, I can't think of any reason why would accrual basis be better than cash, or why would an individual use a fiscal year other than the calendar year.\\ \hline              
             \textbf{Negative answer:} If you didn't receive the money in 2012 or have constructive receipt you really can't claim the income. If the company is going to give you a 1099 for the work they aren't going to give you one until next year and if you claim it this year you will have a hard time explaining the income difference. On the other hand if this isn't miscellaneous income, but rather self employment income and expenses you should be able to claim the expenses in 2012 and if you have a loss that would carry over to 2013. Note it is possible to use an accrual basis if you are running a business (which would allow you to do this), but it is more complex than the cash accounting individual tax payers use.\\ \hline              
\end{tabular}
\captionof{table} {Example of a question, ground truth answer, and negative answer from the FiQA dataset \cite{fiqa}}
\label{tab:QA}
\end{minipage}
\end{center}

\section{Task Definition}
\label{section:task}

QA research can be divided into factoid QA, which extracts facts in the form of text spans from a given passage and non-factoid QA, which retrieves and selects the most relevant sentence or passage level-text such as descriptions, opinions, or explanations as the answer. Since the latter can serve as a basis for applications such as knowledge base construction and information extraction \cite{qa-lstm}, we focus our work on non-factoid answer selection on the passage-level, which is formally defined as: given a question $q$ and an answer candidate pool of size $s$, $\{a_{1}, a_{2}, ..., a_{s}\}$, the goal is to find the best answer candidate, $a_{k}$, for $1 \leq k \leq s$. If the selected answer, $a_{k}$, is inside the set of ground truth answers where one question could have more than one correct answer, the question $q$ is considered to be answered correctly \cite{feng}. The subsequent section will discuss the approach of our work and contributions.

\section{Approach}
\label{section:intro_approach}

We propose a novel QA system by formulating the non-factoid answer selection task into a re-ranking problem that uses techniques from both Information Retrieval (IR) and NLP. Our QA system consists of an Answer Retriever and Answer Re-ranker. The Answer Retriever first returns a list of potentially relevant candidate answers using BM25 \cite{ir}, a simple IR approach, to reduce the size of the answer pool. The Answer Re-ranker then re-ranks and selects the most relevant answers.

The focus of this work is to compare the use of various pre-trained BERT language models for the Answer Re-ranker. We build the models based on different existing learning, further pre-training, and fine-tuning approaches. We examine the effects of further pre-training BERT on a large financial corpus using \cite{araci}'s FinBERT model as well as further pre-training BERT on our own target task corpus. Moreover, we investigate the effectiveness of further fine-tuning using \cite{tanda}'s Transfer and Adapt approach.

The experiments conducted are based on task 2 of the FiQA dataset \cite{fiqa}, which contains opinionated questions and answers crawled from community QA sites, such as Stackexchange, under the Investment topic. Since the corpus is opinion-based, the results of this thesis are also relevant to the application of answer quality prediction in community QA sites.

\begin{figure}[h!]
\centering
  \includegraphics[scale=0.4]{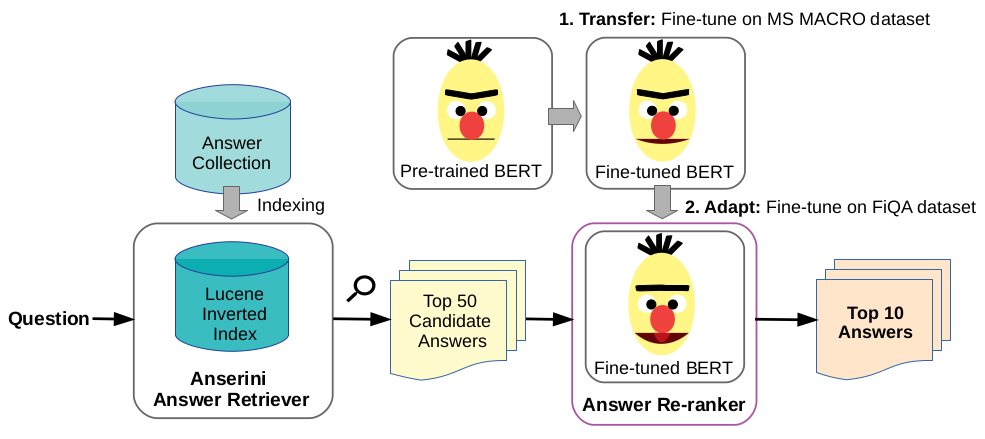}
  \caption{Overview of the QA pipeline. The Anserini Answer Retriever first returns the top 50 candidate answers. The Answer Re-ranker is built by (1) transferring a pre-trained BERT model to the large-scale MS MACRO dataset and (2) adapting the transfered BERT model to the target FiQA dataset. The Answer Re-ranker then outputs the top 10 most relevant answers.}
  \label{fig:QA_pipeline}
\end{figure}

Our experiments show that using the pointwise learning approach to fine-tune a pre-trained BERT model along with the Transfer and Adapt fine-tuning method outperforms the state-of-the-art results significantly. We name this fine-tuned model FinBERT-QA for it's application in financial QA. The overview of the QA pipeline is shown in Figure \ref{fig:QA_pipeline}.

\subsection{Contributions}

To our best knowledge, this is the first time a financial domain QA system using pre-trained BERT language models has been developed and the contributions of this work are as follows \footnote{The code and models for our system can be found here: https://github.com/yuanbit/FinBERT-QA}:

\begin{itemize}
    \item We build FinBERT-QA, which improves the state-of-the-art result of task 2 of the FiQA challenge by 16\% on MRR, 17\% on NDCG, and 21\% on Precision@1.
  \item The pointwise learning approach is shown to be more efficient and effective for fine-tuning BERT for a financial QA task.
  \item We show the effects of further pre-training BERT on a large financial-domain corpus and on the target task corpus are not significant.
  \item The Transfer and Adapt method consisting of multiple fine-tuning steps is shown to be more effective than further pre-training BERT.
  \item Transfer learning with fine-tuned BERT models is shown to outperform traditional IR and representation learning methods.
\end{itemize}

The rest of the paper is structured in the following order: \nameref{chap:background}, \nameref{chap:RelatedWork},  \nameref{chap:Approach}, \nameref{chap:Results}, \nameref{chap:conclusion}.

\chapter{Theoretical Background}\label{chap:background}

We will provide preliminary information in this chapter to understand the most essential NLP and Deep Learning concepts relevant to this work.

\section{Neural Networks}

\subsection{Neural Units}

As a family of classifiers that can deal with non-linear decision boundaries, neural networks are fundamental computational tools for language processing. A neural network consists of neurons which are computational units that take vectors of numerical values as input and produces a single output value \cite{jurafsky}. More specifically, given a set of $n$-dimensional inputs $x_{1}...x_{n}$, a neuron computes a weighted sum $z$ of $x_{i}$, with an additional bias term $b$, which is represented in vector notation as:
\begin{equation}\label{eq:1}
\begin{split}
z & = \sum_{i}^{n} w_{i}x_{i} + b \\
  & = w^{T}x+b
  \end{split}
\end{equation}

$z$ is then passed through an activation function with the sigmoid function as a common choice since it maps the output into the range $[0, 1]$, hence squashing outliers toward 0 or 1. Let $y$ be the output of a neuron passed through a sigmoid activation function:
\begin{equation}
y = \sigma(z) = \frac{1}{1+e^{-z}}\label{eq:3}
\end{equation}

Putting together \eqref{eq:1} and \eqref{eq:3}, the output of a neuron $y$ is:

\begin{equation}
y = \sigma(w^{T}x+b)) = \frac{1}{1 + e^{-(w^{T}x+b)}}
\end{equation}

\subsection{Feedforward Network}

\begin{figure}[h!]
\centering
  \includegraphics[scale=0.5]{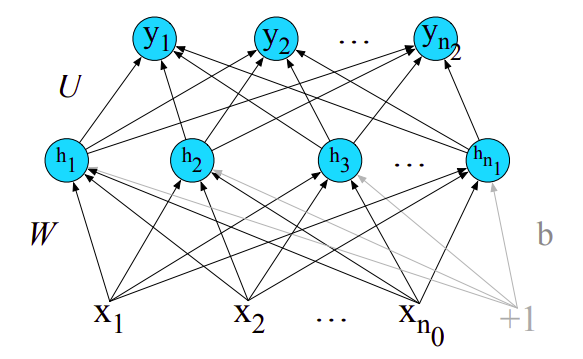}
  \caption{A feedforward network with one hidden and one output layer where $x_{1}...x_{n_{0}}$ is a set of $n$-dimensional inputs, $h_{1}...h_{n_{1}}$ is the output of the hidden layer, and $y_{1}...y_{n_{2}}$ is the result of the output layer. $W$ is the weight matrix of the hidden layer, $U$ is the weight matrix of the output layer, and $b$ is the bias vector \cite{jurafsky}.}
  \label{fig:feedforward}
\end{figure}

The feedforward network shown in Figure \ref{fig:feedforward} is a foundamental neural network, which contains an input, a hidden, and an output layer. The features from the input layer are fed into the hidden layer made up of multiple neurons and the resulting outputs represent a weighted combination of features. Each layer is fully-connected, signifying that there is a link between every pair of input and hidden units \cite{jurafsky}. In a classification task, the output of the hidden layer, a vector of real-values, is fed into a softmax activation function to be converted into a probability distribution of values between 0 and 1. Given a $d$-dimensional vector $z$, the softmax function is defined as:
\begin{equation}\label{eq:4}
softmax(z_{i}) = \frac{e^{z_{i}}}{\sum_{j = 1}^{d}e^{z_{j}}} \; 1 \leq i \leq d
\end{equation}

\subsubsection{Training Neural Nets}

The feedforward network is trained in a supervised fashion by first computing the forward pass, which maps the inputs to an estimated output, $\hat{y}$. Given an input vector $x$, the output of each layer can be represented as weight matrices $W^{[i]}$ and bias vectors $b^{[i]}$ which combines the weight vector and bias in each layer where the superscript represent the $i^{th}$ layer. Let $g(\cdot)$ be the activation function, where the intermediate layers use the sigmoid function and the output layer uses the softmax function. Let $a^{[i]}$ be the output from layer $i$ and $z^{[i]}$ be the combination of weights and biases, the algorithm for computing the forward pass in an $n$-layer feedforward network is:

\hspace*{4mm} \textbf{for} \textit{i} \textbf{in} 1...n\\
\hspace*{8mm} $z^{[i]} = W^{[i]} a^{[i-1]} + b^{[i]}$\\
\hspace*{8mm} $a^{[i]} = g^{[i]}(z^{[i]})$\\
\hspace*{4mm} $\hat{y} = a^{[n]}$

The model learns the parameters $W^{[i]}$ and $b^{[i]}$ for each layer $i$ to optimize the error between the estimate $\hat{y}$ and the true $y$ using backpropagation. Backpropagation uses computational graphs and the chain rule to compute each of the partial derivatives of the loss function from right to left and multiplying the necessary partials to the result in the final derivative. This process is repeated and the number of epochs represent how many times the training algorithm works through the entire dataset \cite{jurafsky}.

Mini-batch training can be used to train on a group of $m$ samples that is less than the whole dataset to increase the efficiency as mini-batches can be vectorized and processed in parallel. The problem of overfitting can arise, where the model attempts to perfectly fit details of the training set and fails to perform well on the test set due to noisy modeling. To avoid overfitting, dropout is a regularization best practice for successful learning, where neurons and their connections are randomly dropped during training to help prevent co-dependency of the neurons that could lead to overfitting \cite{jurafsky}.

\section{Recurrent Neural Networks}

Textual inputs of NLP tasks are sequences that need to be processed through time. Feedforward neural networks are not suitable for text sequences because the input vectors and their corresponding weights are fixed in size, which tend to capture all relevant aspects of a sample at once \cite{jurafsky}.

Recurrent Neural Networks (RNN's) process sequences one input at a time and the hidden layer from the previous time step serve as a memory mechanism for the context. Specifically, the previous hidden layer includes information of the context all the way back to the beginning of the sequence. In order to achieve this, RNN's have an additional set of weights that connect the hidden layer from the previous time step, $h_{t-1}$, to the current hidden layer $h_{t}$. These weights, therefore, provide the model with past context information when computing a current output. RNN's can also be stacked or trained on an input sequence in a backwards order. The result of combining the forward and backward networks is a bidirectional RNN, where the processed inputs from both directions are concatenated to form a single representation that captures both the left and right contexts of an input at each point in time \cite{jurafsky}.

\subsection{Long Short-Term Memory}

In many language applications, distant information within a sequence can be critical. Vanishing gradient problems occur during backpropagation since the hidden layers are required to compute repeated multiplications and the gradients are eventually driven to zero. Long short-term memory (LSTM) networks are used to maintain relevant context over time by training the network to forget (remove) information that is no longer needed and remember (add) information likely to be needed later. To accomplish this, an additional context layer and three gates are added to the architecture. The gates are neural units that operate on the input, previous hidden layer, and previous context layers\cite{jurafsky}.

\section{Encoder-Decoder Models}

\begin{figure}[h!]
\centering
  \includegraphics[scale=0.5]{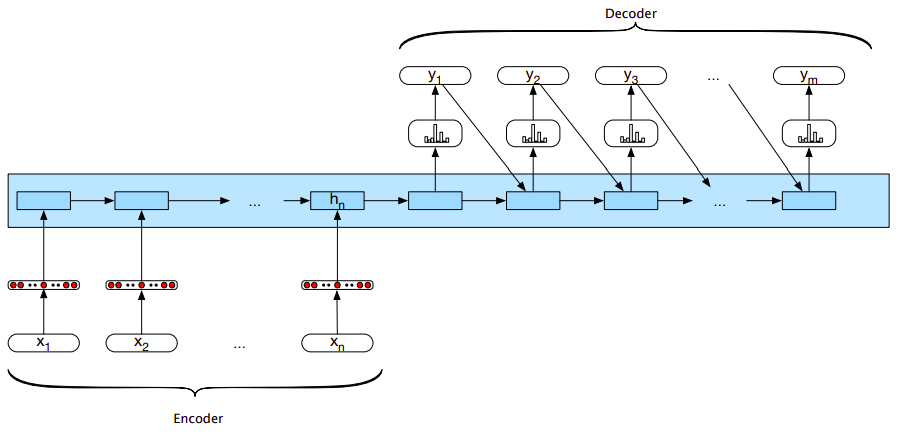}
  \caption{RNN-based encoder-decoder architecture where the encoder passes the final hidden state $h_{n}$ of an RNN into the decoder. The decoder then autoregressively generates a task-specific novel output sequence \cite{jurafsky}.}
  \label{fig:encoder_decoder}
\end{figure}

RNN's can be applied to transduction tasks, where the input sequences are transformed into novel output sequences using sequence-to-sequence models, also known as encoder-decoder models. For example, machine translation and QA are tasks that use encoder-decoder models. The encoder-decoder architecture, typically implemented with RNN's, shown in Figure \ref{fig:encoder_decoder}, consists of an encoder network which passes the final hidden state, $h_{n}$, of an RNN into the decoder network. The decoder then autoregressively generates a task-specific output sequence such as the answer to a question. Autoregressive generation is the process of modeling a novel sequence using the following steps:
\begin{enumerate}
\item The network randomly samples a word from the softmax output resulting from using the sentence marker $\langle s \rangle$ which marks the beginning of the sentence as input.
  \item Then the word embedding of the sampled word is used as the input for the next time step.
  \item Subsequent words are then sampled in the same fashion until the sentence marker $\langle /s \rangle$ which marks the end of the sentence or when the fixed length limit is reached
\end{enumerate}

\subsection{Transformers}

A RNN-based encoder-decoder architecture has a linear sequential nature and does not allow for parallelization within training samples. This becomes critical for sequences with longer lengths since mini-batching across training samples are limited by memory constraints and it is difficult to learn dependencies between distant positions. The Transformer is a model architecture that overcomes the parallelization constraints of RNN's, which relies entirely on an attention mechanism without RNN. Transfromers can learn global dependencies between the input and output significantly faster and achieve better results compared to architectures based on recurrent or convolutional layers \cite{transformers}.

\begin{figure}[h!]
\centering
  \includegraphics[scale=0.3]{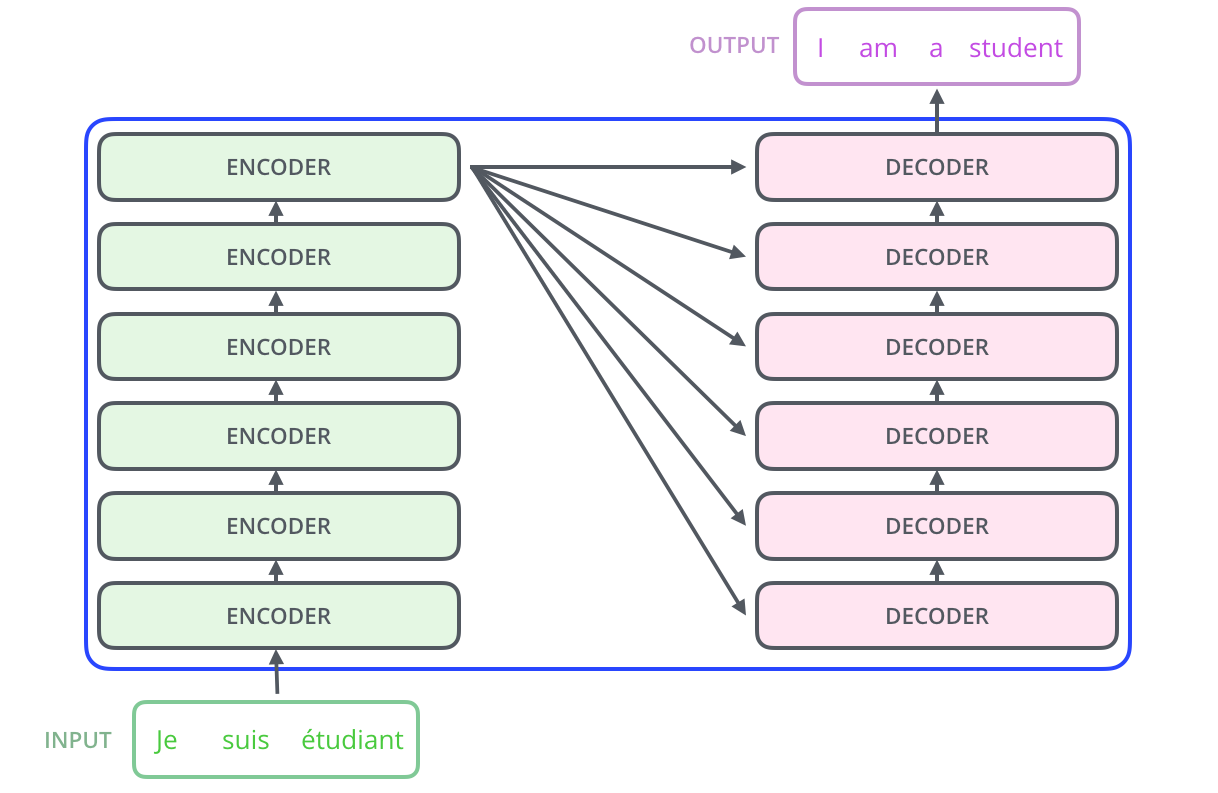}
  \caption{Stacked encoders and decoders of the Transformer architecture for a French to English translation task \cite{alammar}.}
  \label{fig:encoder_decoder_stack}
\end{figure}

\begin{figure}[h!]
\centering
  \includegraphics[scale=0.5]{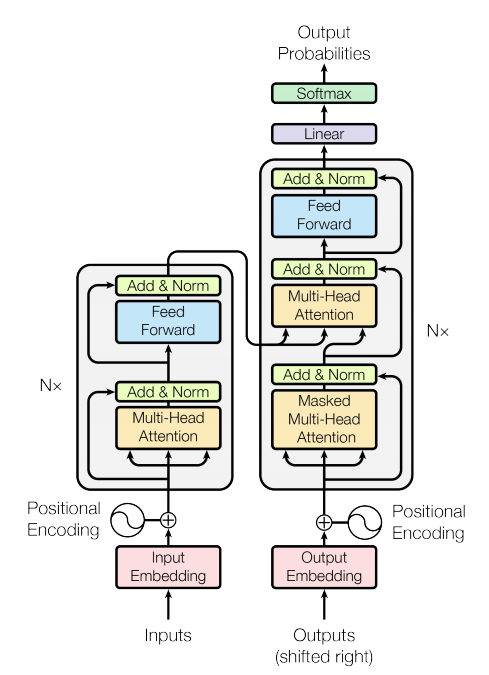}
  \caption{The Transformer model architecture where the left block is the encoder and the right is the decoder. Both the encoder and decoder contain multi-head attention, normalization, and fully-connected feed-forward sublayers. The decoder also contains the masked multi-head attention layer \cite{transformers}.}
  \label{fig:transformers}
\end{figure}

The Transformer architecture is made up of stacks of six encoders and six decoders as shown in Figure \ref{fig:encoder_decoder_stack}. As shown in Figure \ref{fig:transformers}, each encoder and decoder connection is broken down into stacked multi-head self-attention and fully-connected feed-forward sub-layers. Dropout is applied and added to each of the sub-layer inputs for normalization in the Add \& Norm layer.

\vspace{10mm}
The Transformer architecture uses a model dimension of 512 for the inputs and outputs. Specifically, the inputs are WordPieces \cite{wordpiece} of the input tokens, which are sub-word units that deal with rare words in the vocabulary and out-of-vocablary words. For example, "playing" can be broken down into the sub-words, "play" and "ing". Since the Transformer architecture contains no recurrence, positional encodings are created to be summed with the input embeddings to provide information about the order of the words in the input sequence. The intuition of adding positional encodings to the input embeddings is that same words at different positions will have different overall representations \cite{transformers}.

\subsubsection{Self-Attention}

Self-attention is an attention mechanism that allows the model to  determine how much focus to place on other parts of the sequence to encode a word at a position \cite{alammar}. More precisely, self-attention processes each word to compute a representation of the sequence. Thus, each position in the encoder and decoder will attend to all positions of the previous layers. Self-attention is computed by first creating a query, key, and value vector for each word from multiplying the input word embeddings with three matrices to be learned during the training process. The query and key vectors have dimension $d_{k}$ and the value vector has dimension $d_{v}$. The three vectors are the input for the attention function and the output is a weighted sum of the values \cite{transformers}.

The weight is a score that is computed by taking the dot products of the query and all key vectors, dividing each product by $\sqrt{d_{k}}$, and applying the softmax function to obtain the weights of the values \cite{transformers}. The softmax score represents how relevant each word is to the other words in the sequence. Hence, by multiplying each value vector by the softmax score, irrelevant words are drown-out and relevant words are kept intact \cite{alammar}. In practice, the input embeddings are packed into a matrix $X$ where each row corresponds to a word in the input sequence. $Q$, $K$, $V$ are the query, key, and value matrices respectively and the output is a self-attention matrix computed simultaneously as: $$Attention(Q, K, V) = softmax\left(\frac{QK^{T}}{\sqrt{d_{k}}}\right)V$$

\subsubsection{Multi-Head Attention}

\begin{figure}[h!]
\centering
  \includegraphics[scale=0.5]{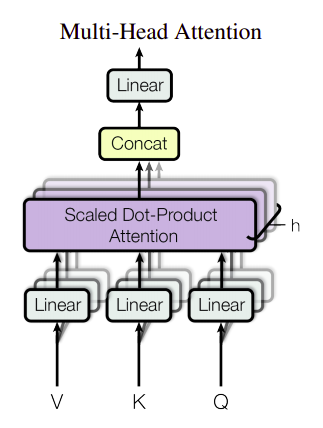}
  \caption{Self-Attention is computed using the Q, query, K, key, and V, value matrices. Multi-Head Attention is the concatenation of mutliple attention heads running in parallel. \cite{transformers}.}
  \label{fig:multi-head-attention}
\end{figure}

Instead of computing a single self-attention function, multi-head attention uses multiple sets of query, key, and value matrices to project the input embeddings into representation subspaces. The attention function is, therefore, performed $h$ times in parallel yielding different output matrices, known as attention-heads. The attention-heads are then concatenated together and multiplied by a weight matix $W^{o}$ to be trained jointly with the model, resulting in the final output matrix that captures information from all attention heads as depicted in Figure \ref{fig:multi-head-attention}. The motivation behind this approach is to expand the model's ability to attend to information from different representation subspaces at different positions \cite{transformers}. 

\begin{figure}[h!]
\centering
  \includegraphics[scale=0.5]{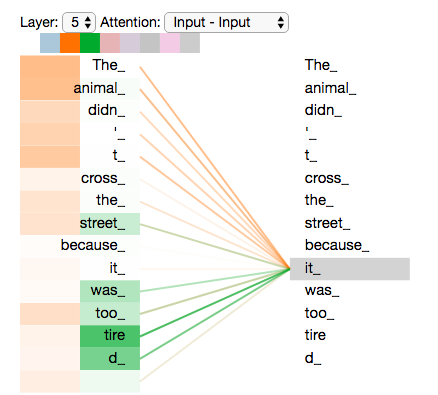}
  \caption{Two attention heads encoding the word "it" represented by the colors beige and green. The intensities of the colors depict the amount of attention. The beige attention head attends "it" to "the animal" while the green attention head attends "it" to "tired" \cite{alammar}.}
  \label{fig:multi_attention}
\end{figure}

For example, the "it" in the sequence "The animal didn't cross the street because it was too tired" could refer to the "animal" or the "street". Figure \ref{fig:multi_attention} shows how the model attends to each word in this sequence using two attention heads. One attention head represented in beige attends "it" to "the animal" while the other attention head represented in green attends "it" to "tired". Therefore the model's representation of "it" takes into account of both "animal" and "tired" as it's representation \cite{alammar}.

\subsubsection{Position-wise Feed-Forward Networks}

After computing multi-head attention and normalizing each of the encoder-decoder layers, the attention output of the word in each position of the sequence is fed into a fully-connected feed-forward layer separately \cite{transformers}. Since the word in each position flows through its own path in the feed-forward layer, there are no dependencies on these paths and the network can represent each position in parallel, overcoming the sequential nature of RNN's \cite{alammar}.

\subsubsection{Decoder}

The decoding phase begins after the normalized multi-head attention and feed-forward sub-layers are processed. Each step of the decoding phase returns an element from the output sequence autoregressively. The architecture of the decoder is similar to the encoder with the difference that the decoder's multi-head attention layer computes an output by creating a query matrix from another attention layer called the masked multi-head attention layer. This layer takes the set of keys and values matrices transformed from the final encoder output as input and  instead of attending to all positions, the masked multi-head attention is only allowed to attend to earlier positions in the output sequence. At the end of the decoding process, the final linear fully-connected neural network projects the decoder output into a logits vector consisting of the relevancy scores of each word in the vocabulary. The logits vector is then passed into the softmax layer turning the scores into probabilities and the word with the highest probability is predicted to be the next-token \cite{alammar}.

\section{Word Embeddings}

To perform NLP tasks, words are represented in a multidimensional semantic space as vectors, also known as embeddings, since the word is embedded in a particular vector space. The context of a word such as neighboring words or grammatical environments, make up a word's distribution and it is assumed that two words occur in similar contexts have similar meanings. Traditionally word embeddings are learned by counting co-occurrences of words \cite{jurafsky}, which can fail to capture contextual relations between words. Instead of count-based approaches, modern neural networks can automatically learn the word embeddings of the input text. The following section will examine two approaches to generate word embeddings: (1) static and (2) contextual.

\subsection{Static Word Embeddings}

Static word embeddings are traditional vector representations, where each word is mapped to a real-valued vector. We will first examine one of the first inspirations of using neural networks to develop static word embeddings, then provide an overview of the two most commonly used embeddings: word2vec \cite{mikolov} and GloVe \cite{glove}.

\subsubsection{Neural Language Models}

Language modeling is a method that assigns a probability to a sequence of words to predict a word from preceding words. The language modeling task is also known as the next-word prediction task. The goal of neural language modeling is to simultaneously learn an embedding for each word in the vocabulary and approximate the probability of a word given the embeddings of the $N$ previous words. An extra embedding layer, also known as the projection layer, is added to the feedforward neural network architecture to accomplish this, where the error is propagated back to the embedding vectors. During training, the one-hot vectors of the $N$ previous words are concatenated together to form the embeddings matrix, which is shared among the $N$ previous words \cite{jurafsky}. 

\subsubsection{word2vec}

\begin{figure}[h!]
\centering
  \includegraphics[scale=0.4]{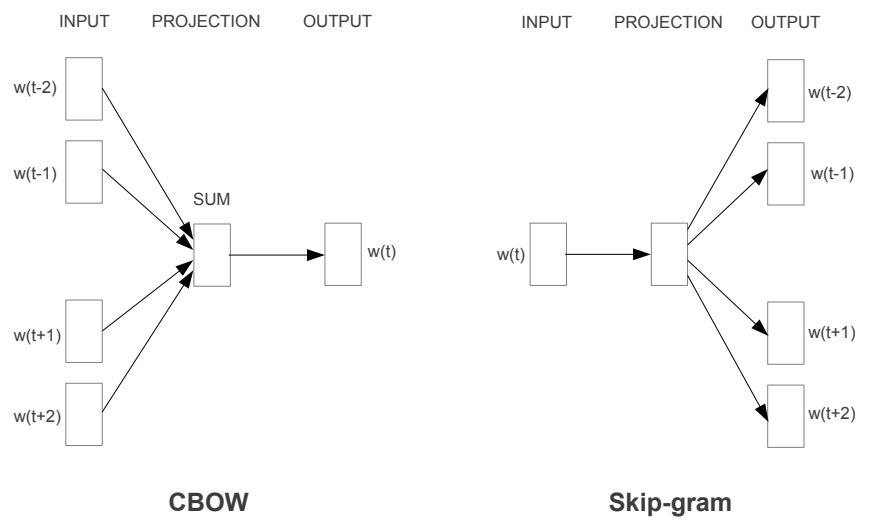}
  \caption{word2vec model architectures. CBOW predicts the center word based on the context and Skip-gram predicts surrounding words given the center word \cite{mikolov}.}
  \label{fig:word2vec}
\end{figure}

Following neural language models, pre-trained word2vec embeddings became widely used to not only represent similar words close to each other, but also answer subtle semantic relationships between words like analogies \cite{mikolov}. Since pre-trained word2vec embeddings were trained on more than 100 billion words and are openly available, they can also be used as the initialization representations of the words for downstream NLP tasks.  As shown in Figure \ref{fig:word2vec}, word2vec introduced two new architectures: (1) Continuous Bag-of-Words (CBOW) and (2) Skip-gram.

The goal of CBOW is to predict a center word using the average of the embeddings of the surrounding context words as input. For example, given the input \{"The", "cat", "over", "the", "puddle"\}, the model tries to predict the center word "jumped". On the other hand, Skip-gram predicts the surrounding context words given a center word. For example, given the input "jumped", the model tries to predict "The", "cat", "over", "the", "puddle" \cite{stanford}. Skip-gram uses negative sampling, meaning that the dataset consists of positive and negative target and context words. Then logistic regression is used to train a classifier to distinguish between the positive and negative samples. The resulting word embeddings are obtained from the regression weights of the classifier \cite{jurafsky}.

\subsubsection{GloVe}

Building on top of word2vec, \cite{glove} proposed a weighted least squares model that captures global corpus statistics based on ratios of probabilities from the word-word co-occurrence matrix. The model outperformed word2vec on several word similarity and NLP tasks. The key difference between GloVe and word2vec is that GloVe estimates the probability of a word occurring in the context of another word based on a global co-occurrence matrix compared to using local context windows of the entire corpus in word2vec. Moreover, instead of using neural networks, GloVe is count-based and uses a least square objective \cite{glove}. Pre-trained GloVe embeddings are also openly available. 

\subsection{Contextual Word Embeddings}

When modeling polysemy, words with multiple meanings, static word embedding methods are suboptimal since they are only allowed to represent each word with a a single vector. As a result, contextual word embeddings derived from pre-trained language models were developed so that each word is assigned a representation that is a function of the entire input sentence. The main difference between contextual and static word embeddings is that contextual embeddings provide not only the embeddings, but also an entire model for the target task. We will examine the three pre-trained language models that achieved breakthroughs in the NLP community: (1) ELMo \cite{elmo}, (2) OpenAI GPT \cite{gpt} (3), and BERT \cite{bert}.

\subsubsection{ELMo}

ELMo \cite{elmo}, Embeddings from Language Models, is one of the first deep contextual word embedding methods that outperformed traditional static word embeddings. ELMo representations are learned from training a two-layer bidirectional LSTM on the language modeling task in a semi-supervised fashion as shown in Figure \ref{fig:bert_comparison}. The model is first pre-trained with unlabeled data to get a representation for each word using unsupervised learning. Then the weights of the language model are fixed and added to a task-specific model for supervised downstream training, which is also known as the fine-tuning step. Instead of just using the top LSTM layer for training, ELMo learns a task-specific linear combination of the vectors stacked above each input word. Thus, the model provides three layers of representations for each input token whereas word2vec and GloVe embeddings only provide one layer of representation \cite{elmo}.

One of the limitations of ELMo is that the context learned during pre-training is only present in the first layer of any downstream model. Moreover, ELMo can't be easily applied to various NLP tasks because the hidden representations from a pre-trained language model are used as additional features for target-task fine-tuning which introduces new parameters for each separate target task.

\subsubsection{OpenAI GPT}

OpenAI's GPT \cite{gpt}, Generative Pre-Training, is a pre-trained language model that outperformed ELMo. It is trained in a semi-supervised manner like ELMo, but instead of a LSTM-based architecture, it uses the Transformer for parallelization and include the language modeling task as an additional task to the fine-tuning step where all of the parameters are fine-tuned using labeled data from the downstream tasks. This improves generalization so that GPT requires minimal changes to the model architecture when adapting to a wide range of NLP tasks.

Even though pre-trained GPT models can be directly fine-tuned on simple tasks like text classification, the models need to be modified for tasks involving inputs with multiple sentence pairs like QA. Moreover, since standard language models are unidirectional, the pre-trained model can only attend to the previous tokens in the self-attention layers of the Transformer as shown in Figure \ref{fig:bert_comparison}. This is suboptimal for tasks such as QA since it is critical to attend to contexts from both directions \cite{bert}. 

\subsubsection{BERT}

BERT \cite{bert}, Bidirectional Encoder Representations from Transformers, is a pre-trained language model that outperformed OpenAI GPT and achieved the state-of-the-art results for eleven NLP tasks in 2018. Solving the limitations of OpenAI GPT, BERT attends to the context bidirectionally and has an unified architecture across different tasks where the fine-tuning step does not require modifying the model for tasks that have multiple sentence input pairs. BERT is pre-trained on a large corpus consisting of the BooksCorpus (800 million words) and English Wikipedia (2,500 million words) and the pre-trained language models are openly available \cite{bert}. In order to achieve the unified architecture, BERT was pre-trained using two unsupervised tasks:

\begin{enumerate}
\item \textbf{Masked Language Modeling (MLM):} The masked language model randomly masks some percentage of the input tokens randomly and the objective is to predict the original vocabulary of the masked word based on its context. This objective requires the language representation to combine both the left and right context, thus, allowing for pre-training a deep bidirectional representation.

\item \textbf{Next Sentence Prediction (NSP):} In order to understand the relationships of two sentence pairs such as QA, a binary classification task is used to predict if sentence B is the next sentence that follows sentence A (labeled as \texttt{isNext} or \texttt{NotNext}). This task allows for a unified architecture for fine-tuning downstream tasks.

\end{enumerate}

\begin{figure}[h!]
\centering
  \includegraphics[scale=0.37]{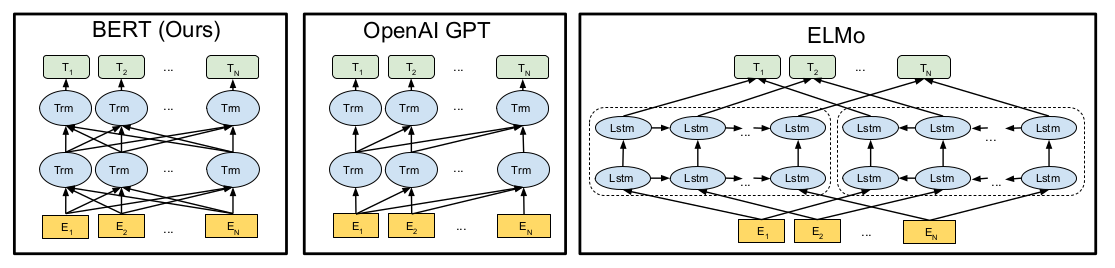}
  \caption{Comparision of pre-training model architectures of BERT, OpenAI GPT, and ELMo, where BERT uses a bidirectional Transformer, OpenAI GPT uses a left-to-right Transformer, and ELMo uses the concatenation of independently trained LSTM's \cite{bert}.}
  \label{fig:bert_comparison}
\end{figure}

BERT's pre-training model architecture is based on the original implementation of the Transformer. Figure \ref{fig:bert_comparison} shows a comparison of the pre-training model architectures of BERT, OpenAI GPT, and ELMo where BERT uses a bidirectional Transformer, OpenAI GPT uses a left-to-right Transformer, and ELMo uses the concatenation of independently trained LSTM's. Thus, BERT is the only model that has representations jointly conditioned on both directions of the context in all layers \cite{bert}.

\begin{figure}[h!]
\centering
  \includegraphics[scale=0.4]{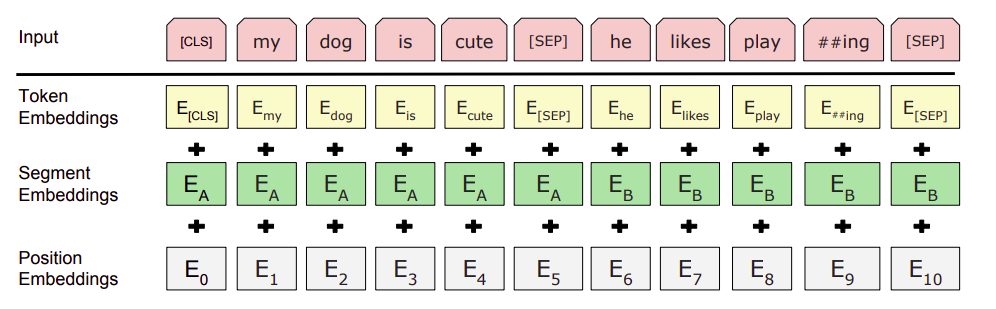}
  \caption{BERT's input representation is the sum of the WordPiece token embeddings, segmentation embeddings, and position embeddings \cite{bert}.}
  \label{fig:bert_input}
\end{figure}

Since BERT has an unified architecture, the input sequence can be both a single sentence or a concatenation of a pair of sentences. WordPiece embeddings with a token vocabulary size of $30,000$ was used as the input embeddings, where the special classification token [CLS] is the first token of every sequence. The special classification token [CLS] encodes the entire input sequence and the final hidden state corresponding to this token is used as the sequence representation in the downstream tasks. If the input sequence is a pair of sentences, the sentences are first separated with a special token [SEP], then additional learned segment embeddings are added to every token indicating whether it belongs to sentence A or B. It follows that, BERT's input representation is the sum of the WordPiece token embeddings, segmentation embeddings, and position embeddings as depicted in Figure \ref{fig:bert_input} \cite{bert}.

Due to BERT's unified architecture, adapting to a downstream task simply involves plugging in the task-specific inputs and outputs and fine-tuning all the parameters end-to-end \cite{bert}, meaning that the training process uses the loss from a downstream task to adjust the weights all the way through the network \cite{jurafsky}. At the output, the final hidden vector corresponidng to the [CLS] token is fed into an output layer for classification tasks \cite{bert}.

\begin{figure}[h!]
\centering
  \includegraphics[scale=0.39]{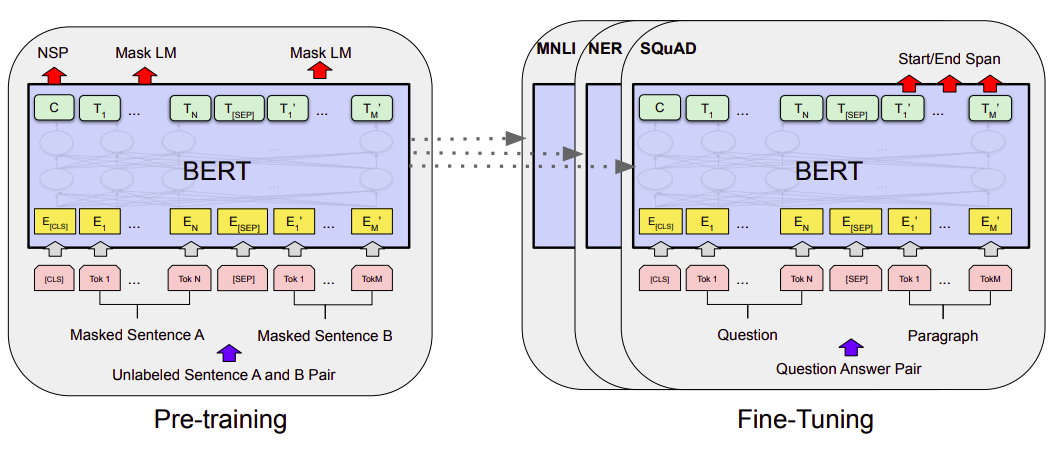}
  \caption{Pre-training and Fine-tuning procedures for BERT where the input sentences are fed into the network as input embeddings, $E$, the final hidden vector of the special token [CLS] is denoted as $C$, and the final hidden vector for the $i^{th}$ input token as $T_{i}$ \cite{bert}.}
  \label{fig:bert}
\end{figure}

Not only is BERT's fine-tuning step simple, it is also relatively inexpensive compared to pre-training. Figure \ref{fig:bert} shows the pre-training and fine-tuning steps of BERT where the input sentences are fed into the network as input embeddings, $E$, the final hidden vector of the special token [CLS] is denoted as $C$, and the final hidden vector for the $i^{th}$ input token as $T_{i}$. The left side shows pre-training using Masked LM and NSP and the right side shows fine-tuning a factiod QA task where sentence A is the question and sentence B is the passage. The right side also indicates that the same fine-tuning procedures can be used for other NLP tasks such as Named-Entity Recognition (NER) and Natural Language Inference (MNLI) \cite{bert}

\chapter{Related Work}\label{chap:RelatedWork}

Initial approaches to QA in the financial domain were focused on rule-based and word-counting systems, which were believed to have good explanatory power \cite{boyer}. Following the expansion of deep learning applications in NLP, neural networks have been applied to QA tasks in attempt to better capture the context in a language corpus. The following section will first discuss techniques that have been used to address the financial QA task and their limitations. We will then examine modern approaches using pre-trained BERT language models for general QA tasks and NLP tasks in the financial domain. The modern approaches will serve as the potential solutions to the financial QA task.

\section{Feature Engineering}

Motivated by the need for advisers of large financial institutions to answer financial outlook and information questions from their clients, \cite{boyer} proposed financial domain QA systems to retrieve the top descriptive text passage answers from a corpus of financial reports given a question. \cite{boyer} first developed a Naive Bayes binary question classifier to determine if a question is a financial outlook or informational question based on financial keyword counts. However, financial entities such as currency, assets, and industry occur more frequently, thus, causing domain term bias and misclassification. To address this issue, a rule-based system was developed where the domain terms were selected and replaced \cite{boyer}. The outlook questions were then treated as factoid questions and the information questions as non-factoid questions. For the non-factoid questions, \cite{boyer} used logistic regression operating over 80 proprietary linguistic scorers.

Another feature-engineeing-based method for a general QA task proposed by \cite{yih} uses the lexical database, WordNet, to pair semantically related words and finding similarities between the questions and answers \cite{qa-lstm}. Since \cite{boyer} and \cite{yih}'s QA systems are based on feature engineering and linguistic match, they could not represent domain-specific financial language \cite{boyer}. Thus, we will examine how deep learning methods can be used to address this problem next.

\section{Representational Learning}

Early deep learning methods for non-factoid QA are based on representation learning, where the goal is to learn a representation vector from the questions and answers and select the most relevant answers by matching the vector representations. Representation learning using neural networks have become widely used in NLP tasks due to it's capability of learning a useful representation for the input text in an unsupervised way, instead of using feature engineering to create representations by hand. The two main representation learning approaches for non-factoid QA are based on the Siamese architecture and the Compare-Aggregate framework.

Since in a non-factiod QA system, each question must be compared with a pool of answer candidates to determine a relevancy score, using the entire answer space as the candidate pool would be inefficient. An efficient way to reduce the answer space is to first employ a non-machine learning answer retrieval system, which focuses on ranking answers that are likely to be relevant. After the retrieval of the candidate answers, an answer re-ranking system can be implemented to re-rank the candidate answers in order to select the most relevant answers. \cite{tran}, for example, proposed such a framework that consists of an answer retriever using BM25 and two answer re-rankers based on the Siamese architecture and Compare-Aggregate framework.

\subsection{Siamese Architecture}

The Siamese architecture applies the same neural network encoder such as RNN's to construct the vector representations of the questions and answers individually. Even though the same encoder is applied, the computation of the representations of the questions and answers are not influenced by each other since they are built separately. \cite{feng} first proposed variations of deep learning frameworks based on Convolution Neural Networks.  \cite{qa-lstm} proposed variations of frameworks to compute the representations of the questions and answers using biLSTM, which are more capable to learning long-range sequential context information. The representations are then connected with a pooling layer in order to use a similarity metric to measure the matching degree of the QA pairs. To achieve better performance, \cite{qa-lstm} introduced variations of the frameworks include building a CNN structure on top of the biLSTM as well as introducing an attention model to the framework.

\cite{tran}'s SRanker is one of the first approaches of applying deep learning to QA in the financial domain using an variation of the Siamese architecture with GloVe embeddings as initializations for the model. However, instead of using a pooling layer, an attention mechanism is applied to give more weight to words that have more influence on the final representation \cite{tran}.

\subsection{Compare-Aggregate Architecture}

The Compare-Aggregate architecture proposed by \cite{wang} first compare small units such as words of the question and answer sequences, then the comparison results are aggregated by a CNN or RNN into a vector representation to calculate the final relevance score. The Compare-Aggregate architecture can capture features between the questions and answers, therefore, it typically has a better performance over the Siamese architecture \cite{lai}. Motivated by \cite{wang}, \cite{tran} used a variation of the Compare-Aggregate architecture to build the CARanker financial non-factiod QA. The CARanker first pre-processes the questions and answers in the Word Representation Layer, then an attention matrix is computed based on the question and answer vectors in the Attention Layer. Next, each answer is matched with a weighted question that best matches the answer in the Comparison Layer by a comparison function. Finally, in the Aggregation Layer, the resulting vector from the previous layer is aggregated using an one-layer CNN to compute the final score that is used to rank the candidate answers \cite{tran}.

\section{Transfer Learning}

Although deep learning methods do not require feature engineering or external knowledge bases, the major downside is that it requires a sufficient amount of labeled training data, which is not always possible especially for domain-specific datasets such as clinical or financial data \cite{tan}. Moreover, representation learning models are based on shallow pre-trained word embeddings such as word2vec and GloVe, which are not able to capture the deep contexts of the dataset. Additionally, domain-specific tasks such as financial QA are especially challenging since labeling financial texts require costly expertise. As a result, transfer learning is a tool to address this problem.

Given a learning task, the goal of transfer learning is to transfer latent knowledge from the source domain with significantly more data to the target domain \cite{tan}. Fine-tuning pre-trained BERT language models is an example of transfer learning, since it was pre-trained on a large source domain with millions of words. In the NLP community, pre-trained BERT language models have been widely used since it achieved the state-of-the-art results for numerous NLP tasks such as non-factoid QA.

\subsection{Fine-tuning BERT for General Non-factoid QA}

\cite{cho} proposed a general QA system by incorporating an answer retriever and re-ranker like \cite{tran}. \cite{cho} first used an answer retriever based on BM25 to retrieve a list of answer candidates. Then instead of using representation learning methods, \cite{cho} used pre-trained BERT models as the answer re-ranker to obtain the most relelvant answer passage texts, which achieved the state-of-the-art results on the MS MARCO passage re-ranking task at the time of it's publication \cite{cho}. The re-ranking task was transformed into a binary classification task with the open-source pre-trained $\text{BERT}_{\textsc{LARGE}}$ model, where the question and answers were concatenated and fed into the network as input for fine-tuning. The fine-tuned model was optimized by a pointwise learning approach, namely the cross-entropy loss. The model outputs a relevance probability for each QA pair and the candidate answers were sorted by this probability to return the top-k relevant answers \cite{cho}. \cite{cho}'s experiments showed that when training on less than $0.3\%$ of the training data (100k QA pairs), the model already outperformed the previous state-of-the-art results by $1.4\%$ on MRR@10. \cite{cho} not only made their fine-tuned models publicly available, but also showed that fine-tuning on pre-trained BERT models require few training samples from the downstream task to achieve good performance \cite{cho}.

\cite{li} also studied fine-tuned BERT models for general non-factoid QA using a similar setup as \cite{cho}. Instead of using a pointwise learning approach, \cite{li} used a pairwise learning approach that takes a pair of correct and incorrect candidate answers and learns which answer is more relevant to the answer. Specifically, given a question, the approach assigns a higher score to correct candidate answers than incorrect candidate answers by a certain margin \cite{li}.

\subsection{Financial Domain Adaption}

Although fine-tuning pre-trained BERT models requires few labeled data to achieve state-of-the-art results for a general non-factiod QA task, the general corpora used to pre-train BERT are not suited for domain-specific downstream tasks such as financial QA since financial texts have specialized language with unique vocabulary. Therefore, further pre-training BERT language models on a financial domain-specific unlabeled corpus can be used as an additional transfer learning step, which allows the model to learn a different semantic relation of the distribution of words in the target downstream task \cite{araci}. In particular \cite{araci} created FinBERT, which further pre-trained BERT models on a large financial corpus. \cite{araci}'s experiments for financial sentiment classification showed that further pre-training a BERT language model on a large financial domain corpus performed the best amongst methods using LSTM and BERT pre-trained on a general corpus. Furthermore, \cite{araci}'s FinBERT achieved the state-of-the-art results on the FiQA task 1 and Financial PhraseBank sentiment scoring tasks \cite{araci}. \cite{araci}'s experiments, thus, show the effectiveness of further pre-training language models on domain-specific corpora.

\subsection{Transfer and Adapt}

Since employing pre-trained BERT models is a relatively new technology, fine-tuning methods are not completely understood and the fine-tuned models can exhibit behaviors such as accuracy variance and always predicting a single label in the binary classification task. As a result, \cite{tanda} proposed a Transfer and Adapt (TANDA) approach which trains Transformer-based models for non-factiod answer selection by applying two fine-tuning steps. The first intermediate fine-tuning transfers the pre-trained language model to a general answer selection dataset that is an order of magnitude larger than the target domain task, then the second subsequent fine-tuning step adapts the transfered model to the target domain task, such as a task with specific types of questions and answers. For example, a general question in the transfer step may be "What is the average heart rate of a healthy person?" and in the adaptation step, the target domain question can have a theme of sport news such as "When did the Philadelphia eagles play the fog bowl?" \cite{tanda}.

It was shown that first using a transferred fine-tuned model greatly reduced the amount of data required for the adaptation step and the fine-tuned model was more stable and robust with low variance of accuracy. In particular, the TANDA approach was applied to fine-tune RoBERTa, a varient of BERT, which improved the state-of-the-art results for the TREC-QA and WikiQA datasets by a significant amount and is ranked first on both leaderboards at the time of this thesis \cite{tanda}.

\chapter{Approach}\label{chap:Approach}

Following the re-ranking approaches used by \cite{tran} and \cite{cho}, the QA system of this thesis consists of two components: (1) Answer Retriever for finding a list of candidate answers and (2) Answer Re-ranker for re-ranking and selecting the most relevant candidate answers. The main focus of this thesis is to compare the use of various pre-trained BERT language models for the Answer Re-ranker and the research objectives are as follows:

\begin{itemize}
\item \textbf{Learning Approach:} There are currently two learning approaches to optimize the loss function when fine-tuning pre-trained BERT language models for non-factoid QA: (1) pointwise and (2) pairwise. Since no comparison has been made to determine which learning approach is more effective for fine-tuning BERT on the non-factoid QA task in the financial domain, we aim to compare the two approaches.
\item \textbf{Further Pre-training:} Despite being a powerful method, BERT is pre-trained on a general corpus from Wikipedia and the BooksCorpus. Since financial passages can be extremely technical, there is a possibility that BERT will underperform when encountering financial languages. As \cite{araci} showed that further pre-training BERT on a large financial corpus achieved better results for financial sentiment analysis, we intend to compare the results of further pre-training BERT for financial non-factoid QA.
\item \textbf{Further Fine-tuning:} \cite{tanda} showed that further fine-tuning BERT using the Transfer and Adapt approach can improve the stability of the downstream model. We will use this approach to first transfer a pre-trained BERT model to a large general non-factoid QA task, then adapt the transfered model to a financial non-factoid QA task. Our goal is to compare the effectiveness of further fine-tuning and further pre-training BERT.
\end{itemize}

\section{Experimental Setup}

In order to setup our experiments we formulate our research objectives into the following research questions (RQ):

\begin{itemize}
  \item \textbf{RQ1}: Which learning approach is more effective for fine-tuning a pre-trained BERT model for the financial non-factoid QA task?
  \item \textbf{RQ2:} What are the effects of further pre-training BERT on a large financial-domain corpus and a target-domain financial QA corpus?
  \item \textbf{RQ3:} How does the Transfer and Adapt further fine-tuning approach compare with further pre-trained BERT models?
  \item \textbf{RQ4}: How do baseline IR and representation learning models compare with fine-tuned BERT models?
\end{itemize}

Our experiments are based on a QA system that uses Anserini's BM25 implementation as our Answer Retriever. For the Answer Re-ranker, we first implement a baseline model using representation learning. Then we develop advanced re-ranking systems based on fine-tuning pre-trained BERT language models. The models for the Answer Re-ranker were implemented using PyTorch and Huggingface's Transformer library \cite{huggingface}. We use a GCP n1-highmem-2 instance with one NVIDIA P100 GPU and 25 GB RAM.

\section{Dataset}

FiQA \cite{fiqa}, Financial Opinion Mining and Question Answering, is an open challenge of the 2018 International World Wide Web Conference with two tasks. We use the opinion-based financial QA dataset from task 2 of the FiQA challenge, which we denote as the FiQA dataset in this work. The goal of the task is to build a QA system that answers questions from a corpus of financial documents in English that came from data sources such as microblogs, reports, and news.

The FiQA dataset was built by crawling Stackexchange, Reddit, and StockTwits posts under the investment topic in the period between 2009 and 2017. The public dataset contains $6,648$ questions and 57,640 answer passages with 17,110 QA pairs \cite{fiqa}. We pre-process the raw dataset to remove special characters, empty answer passages, and QA pairs containing empty answers. There is at least one answer for each question and further analysis of the dataset is as follows:

\begin{itemize}
  \setlength\itemsep{-1em}
  \item \textbf{Average number of answers for each question:} 3
  \item \textbf{Maximum number of answers for each question:} 23
  \item \textbf{Average Question length:} 11
  \item \textbf{Average Answer length:} 136
\end{itemize}

The pre-processed dataset has 57,600 answers. Following \cite{tran}'s experimental setup we split the questions into training, validation, and test sets. If the entire answer space of 57,600 unique answers are used, then each question must be compared with each of the answers. Due to limited computing resources, we use an Answer Retriever to first return the top 50 candidate answers, reducing the answer pool size. 

We generate a list of negative answers for each question by subtracting the 50  candidate answers from the list of ground truth answers. We then create our training, validation, and test samples, where each sample is a list of triples with a question id, a list of ground truth answer ids, and a list of negative answer ids. The number of questions, QA pairs, and samples for each dataset is shown in Table \ref{tab:data}.

\begin{center}
 \begin{tabular}{||c c c c||}
 \hline
  & \textbf{\# Questions} & \textbf{\# QA pairs} & \textbf{\# Samples}\\
 \hline
 \textbf{Training} & 5,683 & 14,721 & 283,707 \\
 \hline
 \textbf{Validation} & 632 & 1,570 & 31,582 \\
 \hline
 \textbf{Test} & 333 & 819 & 16,500 \\
 \hline
\end{tabular}
\captionof{table}{Number of questions, QA pairs, and samples for each dataset}
\label{tab:data}
\end{center}

\section{Models}

\subsection{Baseline Systems}

We use the Answer Retriever based on BM25 and our implementation of QA-LSTM \cite{qa-lstm}, a basic representation learning model as our baseline Answer Re-rankers.

\subsubsection{BM25}
Our first baseline system, BM25, is an IR approach that uses a simple weighted inverted index system based on TF-IDF. TF-IDF computes the product of a term's frequency (TF) and the inverse document frequency (IDF) with the intent to give an importance weighting for terms. For example, if a term occurs in a high frequency in many documents, it would be given a lower score. On the other hand, if a term appears many times within a small number of documents, it would be given a higher score \cite{ir}.

Building on top of TF-IDF and taking the document lengths into account, the resulting value used for ranking the relevance of documents given a query is the Retrieval Status Value (RSV):
$$ RSV_{d} = \sum_{t \in q} log \left( \frac{N}{df_{t}} \right)  \frac{(k_{1} + 1)tf_{td}}{k_{1}((1-b)+b(L_{d}/L_{ave})) + tf_{td}}$$

where $tf_{td}$ is the frequency of term $t$ in document $d$, and $L_{d}$ and $L_{ave}$ are the length of the document $d$ and the average document length for the whole collection. $k_{1}$ and $b$ are tuning parameters, where $k_{1}$ adjusts the scaling of the document term frequency. For example, a $k_{1}$ of value of $0$ corresponds to a model with no term frequency, and a large value corresponds to using raw term frequency. $b$ is between $0$ and $1$ and it determines the scaling by the document length, where if $b = 1$, then the term weight is fully scaled by the document length. If $b=0$, then there is no length normalization \cite{ir}.

Our Answer Retriever uses Anserini's \cite{anserini} BM25 implementation to retrieve $50$ candidate answers for each question. Anserini is an open-source IR toolkit built on top of Lucene. We use the list of answers from the FiQA dataset to first build an inverted index, then we use Pyserini, a Python interface of Anserini, to generate the candidate answers. We use Anserini's default BM25 tuning parameter values, $k{1}=0.82$ and $b=0.68$. 

\subsubsection{QA-LSTM}

\begin{figure}[h!]
\centering
  \includegraphics[scale=0.5]{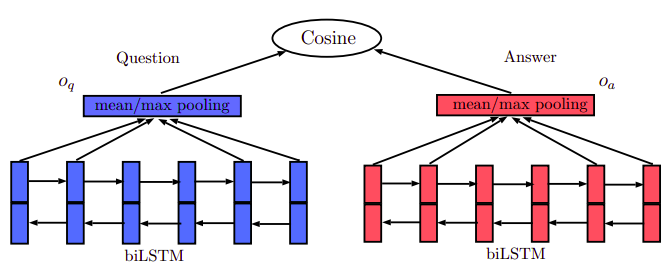}
  \caption{QA-LSTM Architecture. The questions and answers are encoded using a shared biLSTM, then a pooling layer is used to generate a single vector representation for both the questions and answers. The cosine similarity is then used to measure the distance between the question and answer representations \cite{qa-lstm}.}
  \label{fig:qa-lstm}
\end{figure}

Our second baseline system, QA-LSTM \cite{qa-lstm}, is a traditional representational learning model that uses the Siamese architecture. QA-LSTM first generates a distributed representation for both the question and answer texts independently using a shared biLSTM encoder. Since the biLSTM outputs are based on the word-level inputs at each time step, a pooling layer is used to generate a single vector representation for both the questions and answers. After applying dropout, the cosine similarity is used to measure the distance between the question and answer representations as depicted in Figure \ref{fig:qa-lstm} \cite{qa-lstm}. 

The implementation details are as follows:

\begin{itemize}
\item \textbf{Input:} After pre-processing, we use Python's NLTK library to tokenize the questions and answers and build a vocabulary of size $85,034$. We then vectorize the questions and answers with the corresponding vocabulary ids. Then, we construct our input for the model by truncating or padding each question and answer vector so that the maximum sequence length is $128$. Our final input into the model is a list of lists with the triples: question id, a list of positive answers, and a list of negative answers.
\item \textbf{Forward Pass:} The question and answer vectors share the same network parameters. We use torchtext's pre-trained GloVe word embeddings, which has 6 billion tokens to initialize the embedding layer with a dimension of $100$. The embedding layer is fed into one layer of biLSTM with a hidden size of $256$. Then we use max pooling to generate a single vector representation for the questions and answers. We use max pooling because \cite{qa-lstm}'s experiments on different pooling strategies showed that max pooling had the best performance for QA-LSTM. We then apply a dropout rate of $0.2$ and use the cosine similarity to compute the measure of relevance between the questions and answers.
\item \textbf{Loss Function:} The training objective is the hinge loss defined as follows:
$$L = max\{0, M - cosine(q, a_{+}) + cosine(q, a_{-})\}$$ 
where $q$ is a question, $a_{+}$ is a ground truth positive answer, $a_{-}$ is an incorrect negative answer, and $M$ is a constant margin. We use a margin of $0.2$. The intuition behind the hinge loss is that when $L$ is positive, the positive answer is ranked below the negative answer. Whereas, if the positive answer has a higher score than the negative answer by as least $M$, then the loss is zero. We treat each question with more than one ground truth answer as multiple training samples \cite{qa-lstm}.
\item \textbf{Training:} We train our model using mini-batch gradient descent for $3$ epochs with a batch size of $64$ and a learning rate of $1e\text{-}3$. We evaluate the test set by using the model with the minimal validation loss.
\end{itemize}

\subsection{Advanced Systems}

\begin{figure}[h!]
\centering
  \includegraphics[scale=0.4]{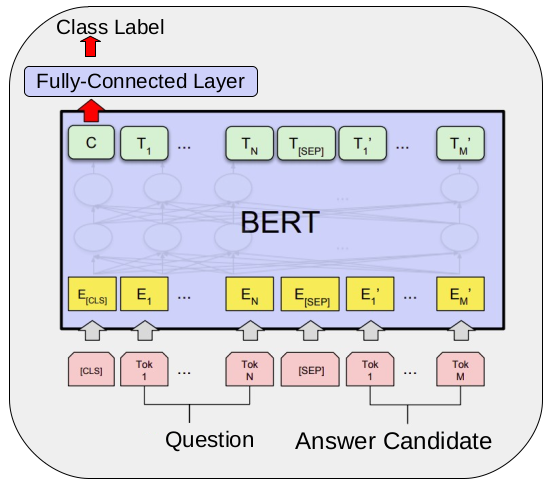}
  \caption{Fine-tuning BERT to classify if the candidate answers are relevant to the question. The output is a probability of the relevance score for each candidate answer.}
  \label{fig:bert_qa}
\end{figure}

For our advanced Answer Re-ranker, we develop 5 variants of fine-tuned BERT models: BERT-Pointwise, BERT-Pairwise, FinBERT-Domain, FinBERT-Task, FinBERT-QA. We use the smaller of the two openly available pre-trained BERT models: $\text{BERT}_{\textsc{BASE}}$ which has 12 encoder layers, a hidden size of 768, 12 multi-head attention heads and 110 million parameters. Following \cite{cho}, we transform the financial QA task into a binary classification task for fine-tuning BERT. We use the [CLS] vector as the input to a fully-connected layer to compute the relevance score of an answer passage with respect to the question. Then we apply the softmax function to each of the output logits to get a probability. We compute this probability for each of the candidate answers given a question, then we sort them in descending order to get the top most relevant answers. The fine-tuning step is shown in Figure \ref{fig:bert_qa}.

The general implementation details are as follows:

\textbf{Input:} After applying the BERT Tokenizer from the Transformer library to the questions and answers, we concatenate the question and each of the answer candidates as one sequence. Then we add the [CLS] and [SEP] tokens to the sequences. The question serves as sentence A and the answer passage as sentence B for the segment embeddings. The first token of every sequence is always a special classification token [CLS], where the final hidden state corresponding to this token is used as the sequence representation for the classification task. The input sequence tokens are padded or truncated to the maximum input sequence length. However, as $\text{BERT}_{\textsc{BASE}}$ uses a maximum input sequence length of 512, the input sequences can only have at most 512 tokens. The BERT Tokenizer returns lists of token embeddings corresponding to the WordPieces in the sequence, lists of segment embeddings of 0's and 1's to distinguish the question from the answer, and lists of attention mask ids indicating which position is part of the sequence and which is a padding. In summary, the following information are the inputs for fine-tuning $\text{BERT}_{\textsc{BASE}}$ for our financial QA downstream task:
\begin{itemize}
\setlength\itemsep{-1em}
\item Token embeddings
\item Segment embeddings
\item Attention Mask ids
\item Binary labels
\end{itemize}

\textbf{Loss Function:} We first experiment with the learning approach using the pointwise cross-entropy loss and \cite{li}'s pairwise loss by fine-tuning $\text{BERT}_{\textsc{BASE}}$ on our downstream task. Since the pairwise approach needs to compute the loss based on both the positive and negative QA pairs, it is more expensive than the pointwise approach. Therefore, for the experiments on further pre-training and fine-tuning BERT, we used the cross-entropy loss for better efficiency.

\textbf{Training:} We train our model in mini-batches using the Adam optimizer \cite{adam}, which is based on adaptive learning rates. Adam, derived from adaptive moment estimation, is an optimizer that computes individual adaptive learning rates for different parameters from estimates of the first and second moments of the gradients \cite{adam}. We use a weight decay of $0.01$ and $10,000$ warmup steps to give the model time for the learning rate to adapt to the data. We select the model with the minimal validation loss for testing.

\textbf{Hyperparameters:} We tune the hyperparamters according to the recommendations of \cite{bert} as follows: The mini-batch size is tuned amongst $\{16, 32\}$. The learning rate is tuned amongst $\{3e\text{-}5, 3e\text{-}6, 3e\text{-}7\}$. The maximum sequence length is tuned amongst $\{128, 256, 512\}$ tokens, and the number of epochs is tuned amongst $\{2, 3, 4\}$. Due to limited resources, we use a maximum sequence length of 128 to compare the learning approaches for more efficient fine-tuning.

We separate the descriptions of our fine-tuned BERT models into three categories: (1) General BERT, (2) Further pre-trained BERT, and (3) Further fine-tuned BERT.

\subsubsection{General BERT}

We develop the following models using $\text{BERT}_{\textsc{BASE}}$ without any further pre-training or fine-tuning:

\textbf{BERT-Pointwise:} We use a pointwise learning approach to fine-tune $\text{BERT}_{\textsc{BASE}}$. The training inputs are triples $(q_{i}, c_{ij}, y_{ij})$, where $q_{i}$ is a question, $c_{ij}$ is a candidate, and $y_{ij}$ is a binary value indicating whether $c_{ij}$ contains the correct answer to $q_{i}$. The binary classifier is trained with the objective to minimize the cross-entropy loss of the question and candidate answer defined as follows:
$$ L = -\sum_{j \in J_{pos}}log(s_{j}) - \sum_{j \in J_{neg}}log(1-s_{j})$$ where $s_{j}$ is the estimated score of how relevant a candidate answer is, $J_{pos}$ is the  set of indexes of the relevant answer passages, and $J_{neg}$ is the set of indexes of non-relevant answers passages in the top 50 answers retrieved with BM25 \cite{cho}. The top-ranked candidate answer of the trained classifier is then selected as the answer during inference \cite{lai}.

\textbf{BERT-Pairwise:} We use the pairwise learning approach proposed by \cite{li}, which assigns a higher score to correct candidate answers than incorrect candidate answers by a certain margin. Specifically, given a question $q$, a positive answer $p$ and a negative answer $n$, the training inputs are triples $(q, p, n)$. During fine-tuning, the triples are split into $(q, p)$ and $(q, n)$ to get the [CLS] embeddings respectively. The training objective is a combination of maximizing the negative cross-entropy of positive and negative samples and minimizing the hinge loss function as follows:
$$L = \lambda_{1}(log\hat{y}_{\theta}(q,p) + log(1-\hat{y}_{\theta}(q,n))) + \lambda_{2}max\{0,m-\hat{y}_{\theta}(q,p) + \hat{y}_{\theta}(q,n)\}$$ where $\hat{y}_{\theta}(q,p))$, $\hat{y}_{\theta}(q,n))$ denote the predicted scores of positive and negative answers, $\lambda_{1} = 0.5$ and $\lambda_{2} = 0.5$ are weighted parameters used in \cite{li}, and $m = 0.2$ is the constant margin for the hinge loss.

\subsubsection{Further Pre-trained BERT}

The following models are further pre-trained on top of $\text{BERT}_{\textsc{BASE}}$ using the run\underline{\hspace{2mm}}language\underline{\hspace{2mm}}modeling.py script from the Transformer library before fine-tuning on the target FiQA dataset.

\begin{figure}[h!]
\centering
  \includegraphics[scale=0.33]{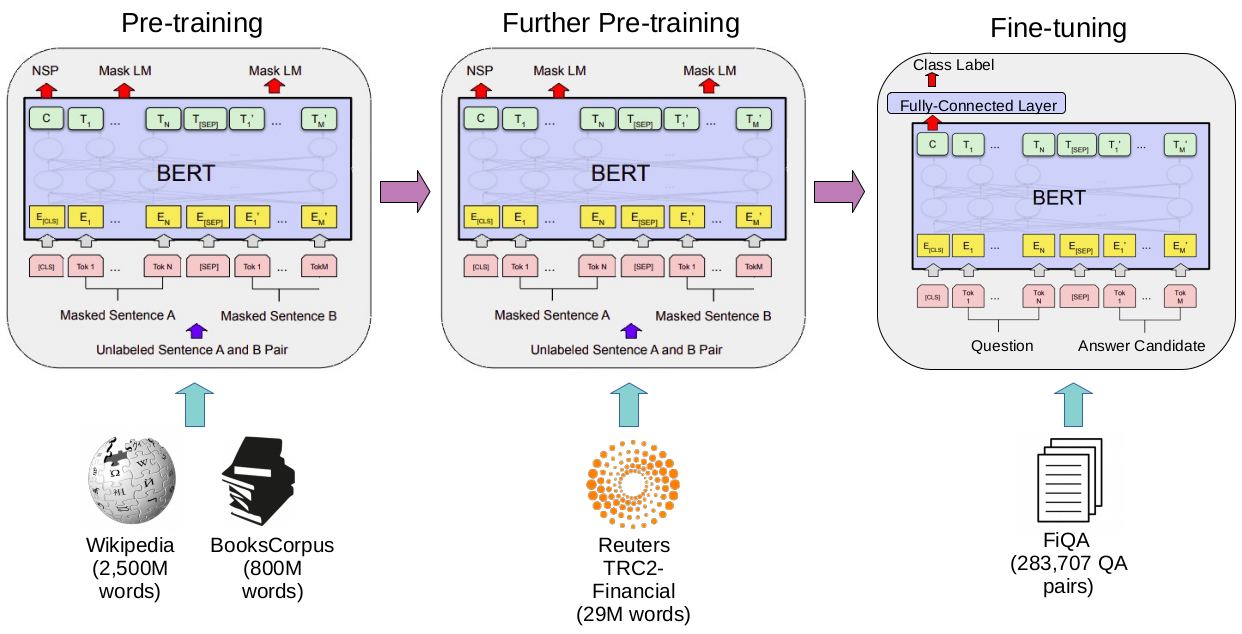}
  \caption{The FinBERT-Domain model further pre-trained $\text{BERT}_{\textsc{BASE}}$ on Reuters TRC2-financial, a large financial domain corpus, before fine-tuning it on the target FiQA dataset.}
  \label{fig:finbert_domain}
\end{figure}

\textbf{FinBERT-Domain:}
We used \cite{araci}'s FinBERT model, which further pre-trained $\text{BERT}_{\textsc{BASE}}$ on Reuters TRC2-financial, a large financial domain corpus, which consists of 46,143 documents with more than 29 million words. Reuters TRC2-financial is a subset of Reuters' TRC24 constructed by \cite{araci} by filtering relevant financial keywords. Like \cite{araci}, we denote this model as FinBERT-Domain. The further pre-training and fine-tuning process is shown in Figure \ref{fig:finbert_domain}.

\begin{figure}[h!]
\centering
  \includegraphics[scale=0.33]{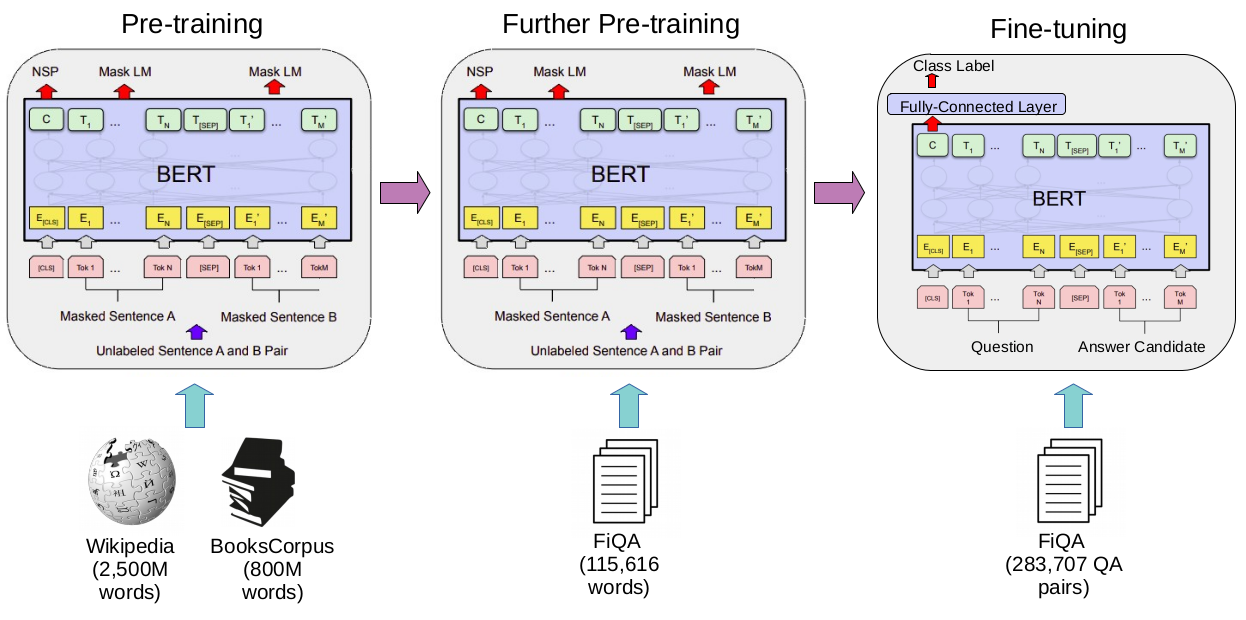}
  \caption{The FinBERT-Task model further pre-trained $\text{BERT}_{\textsc{BASE}}$ on the target FiQA dataset before fine-tuning.}
  \label{fig:finbert_task}
\end{figure}

\textbf{FinBERT-Task:} We further pre-trained $\text{BERT}_{\textsc{BASE}}$ using data directly from our downstream FiQA task dataset, which consists of 115,616 words as shown in Figure \ref{fig:finbert_task}. Like \cite{araci}, we will also denote this model by FinBERT-Task. We further pre-trained the model for 1 epoch with a batch size of 8 and maximum sequence length of 512.

\subsubsection{Further Fine-tuned BERT}

The following model applies \cite{tanda}'s Transfer and Adapt two-step further fine-tuning approach.

\textbf{FinBERT-QA:} We want to first transfer $\text{BERT}_{\textsc{BASE}}$ to a general QA task with a dataset that is much larger than the FiQA dataset. Due to limited computing resources, we use \cite{cho}'s open-source model as the transferred model of our first fine-tuning step. \cite{cho}'s model fine-tuned $\text{BERT}_{\textsc{BASE}}$ on the MS MARCO Passage Ranking task \cite{ms}. MS MARCO, Microsoft Machine Reading Comprehension, is a large-scale dataset focused on machine reading comprehension, question answering, and passage ranking. The dataset of the Passage Ranking task contains 1 million queries from real users and their respective annotated relevant answer passages. Since each query is paired with the top 1,000 answers retrieved from BM25, the training set contains approximately 400 million tuples of query, relevant, and non-relevant answer passages. Table \ref{tab:MS} shows an example of a question and a ground truth answer from the MS MACRO Passage Ranking dataset.

\begin{center}
\begin{minipage}{\linewidth}
\centering
\renewcommand{\arraystretch}{1.5}
\begin{tabular}{||>{\arraybackslash}p{12cm}||}

            \hline
            \textbf{Question:} what is the nutritional value of oatmeal?\\ \hline
             \textbf{Ground truth answer:} Oats make an easy, balanced breakfast. One cup of cooked oatmeal contains about 150 calories, four grams of fiber (about half soluble and half insoluble), and six grams of protein. To boost protein further, my favorite way to eat oatmeal is with a swirl of almond butter nestled within.\\ \hline              
\end{tabular}
\captionof{table} {Example of a QA pair from the MS MACRO Passage Ranking dataset \cite{ms}}
\label{tab:MS}
\end{minipage}
\end{center}

\begin{figure}[h!]
\centering
  \includegraphics[scale=0.33]{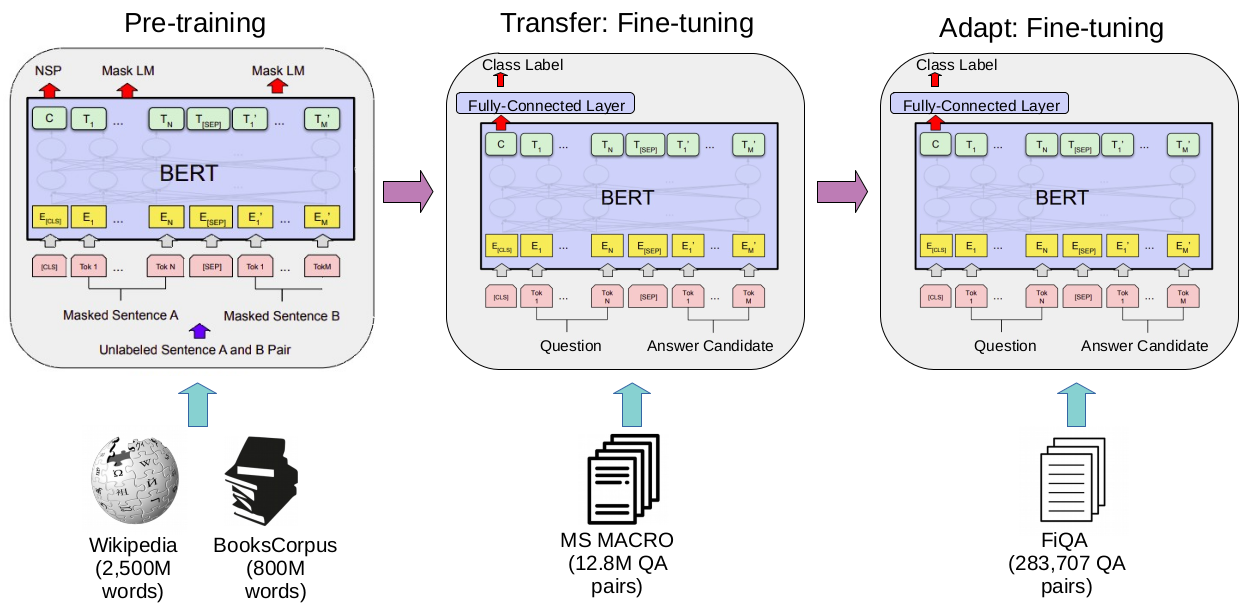}
  \caption{The FinBERT-QA model uses the two-step Transfer and Adapt fine-tuning approach. $\text{BERT}_{\textsc{BASE}}$ was first transferred to and fine-tuned on the large-scale MS MACRO dataset then the transferred model was adapted to and fine-tuned on the target FiQA dataset.}
  \label{fig:tanda}
\end{figure}

\cite{cho}'s model was trained on a subset of the MS MACRO Passage Ranking task, which contains 12.8 million QA pairs. \cite{cho} found that fine-tuning the model using less than $2\%$ of the full training set was sufficient and there was no further improvement when training for another 3 days, which is equivalent to 50 million QA pairs. We convert \cite{cho}'s publicly available TensorFlow model to a PyTorch model using Huggingface's Transformer library, then adapt the model using a subsequent fine-tuning step on the target FiQA dataset as show in Figure \ref{fig:tanda}. We name this model FinBERT-QA since it transfers a pre-trained BERT model to a general QA task then adapts the transfered model to the financial domain.

\section{Evaluation Metrics}

To evaluate our systems, we use a test set derived from the FiQA training set. We use the Mean Reciprocal Rank (MRR) and Normalized Discounted Cumulative Gain (NDCG) for the top 10 answers as well as Precision@1 to evaluate the systems' performances.

\subsection{Mean Reciprocal Rank}

The Mean Reciprocal Rank (MRR) is the mean of the Reciprocal Rank (RR) score across multiple queries. The RR is defined as $\frac{1}{k}$  where $k$ is the rank position of the first relevant ground truth answer \cite{stanford2}.

\subsection{Normalized Discounted Cumulative Gain}

The Normalized Discounted Cumulative Gain (NDCG) is a particular measure of the Cumulative Gain (CG). The CG is the sum of the relevance scores of the top $k$ retrieved documents. However, if the relevance scores are non-binary, then a highly relevant document with a lower rank will not change the results of the CG. As a result the Discounted Cumulative Gain (DCG) is used to make sure a more relevant document is discounted if it has a lower rank. DCG is defined as follows:
$$DCG@k = rel_{1} + \sum_{i=2}^{k} \frac{rel_{i}}{log_{2}(i+1)}$$ where $k$ is the top $k$ retrieved documents and $rel_{i}$ is the relevance score at position $i$ \cite{stanford2}. In order to normalize larger result lists, NDCG normalizes DCG at rank $k$ by the DCG value of the ideal ranking (iDCG). The ideal ranking is a result list where the relevance scores of the documents are ranked in descending order \cite{stanford2}. NDCG is therefore defined as: $$nDCG@k = \frac{DCG@k}{iDCG@k}$$

\subsection{Precision}

Precision is a measure of the return of true positives. We use Precision@1 to determine the percentage of retrieved documents that are relevant to the query at the top 1 position \cite{ir}. Precision is defined as:
$$ Precision = \frac{\#(relevant\;items\;retrieved)}{\#(retrieved\;items)} = \frac{true\;positives}{true\;positives + false\;positives}$$

\chapter{Experimental Results}
\label{chap:Results}

We find that training our advanced models for 3 epochs with a batch size of 16, a learning rate of $3e\text{-}6$, and a maximum sequence length of 512 was the most effective for fine-tuning BERT on the financial QA task. In fact, the models have great performance when just training for 2 epochs. 

We compare our results with the results from \cite{fiqa} and \cite{tran}. The results from \cite{fiqa} used the official test set from the FiQA task 2 challenge while \cite{tran} used a subset of the training set. Therefore, the models from \cite{fiqa} have an advantage over \cite{tran} and our models since they could use the entire training set to train their models, while we are constricted to using a subset. Even so, \cite{tran}'s CARanker model built with the Compare-Aggregate architecture outperformed the models from \cite{fiqa} and had the state-of-the-art (SOTA) results. 

Since there are no further information about the models from \cite{fiqa}, we cannot compare our approach directly to theirs. Therefore, we focus on comparing our results from the fine-tuned BERT models with \cite{tran}'s models as well as with our baseline BM25 and QA-LSTM models.

\begin{center}
 \begin{tabular}{||c c c c||}
 \hline
  \textbf{Model} & \textbf{MRR@10} & \textbf{NDCG@10} & \textbf{Precision@1}\\
 \hline
  CUKG\underline{\hspace{2mm}}Tongji \cite{fiqa} & 0.096 & 0.168 & - \\
 eLabour \cite{fiqa} & 0.195 & 0.305 & - \\
 \hline
 SRanker$_{mlp}$ \cite{tran} & 0.242 & 0.278 & 0.119 \\
 SRanker$_{bilinear}$ \cite{tran} & 0.268 & 0.297 & 0.153 \\
 CARanker (SOTA) \cite{tran} & 0.279 & 0.308 & 0.157 \\
 \hline
 BM25 & 0.305 & 0.361 & 0.228 \\
 QA-LSTM & 0.098 & 0.143 & 0.054 \\
 \hline
 BERT-Pointwise & 0.433 & 0.473 & 0.366 \\
 \textbf{FinBERT-Domain} & 0.433 & 0.463 & \textbf{0.375}\\
  FinBERT-Task & 0.435 & 0.477 & 0.366 \\
  \textbf{FinBERT-QA} & \textbf{0.436} & \textbf{0.482} & 0.366 \\
 \hline
\end{tabular}
\captionof{table}{Experimental results based on the FiQA dataset. The results of the models from \cite{fiqa} are reported in the  first group. The results of \cite{tran}'s representation learning-based results are reported in the second group. The baseline BM25 and our QA-LSTM implementation results are reported in the third group. The results from the fine-tuned BERT models are reported in the fourth group.}
\label{tab:result}
\end{center}

\section{Learning Approach (RQ1)} 

As shown in Table \ref{tab:learning}, the pointwise learning approach has slightly better results compared to the pairwise learning approach. This may be because the QA task was transformed into a binary classification problem, therefore, the cross-entropy loss works better. Even though the differences are minimal, the pointwise approach is more efficient since the pairwise approach needs to compute the loss based on both the positive and negative QA pairs.

\begin{center}
 \begin{tabular}{||c c c c||}
 \hline
  \textbf{Model} & \textbf{MRR@10} & \textbf{NDCG@10} & \textbf{Precision@1}\\
 \hline
 \textbf{BERT-Pointwise} & \textbf{0.366} & \textbf{0.416} & \textbf{0.291} \\
 \hline
 BERT-Pairwise & 0.347 & 0.406 & 0.270 \\
 \hline
\end{tabular}
\captionof{table}{Comparison of the pointwise and pairwise learning approaches for fine-tuning BERT on the financial QA task}
\label{tab:learning}
\end{center}

\section{Further Pre-training (RQ2)} 

The results in Table \ref{tab:result} show that further pre-training BERT on a large financial domain corpus only has a slightly better Precision@1. The results for the MRR are the same and the NDCG did slightly worse. On the other hand, further pre-training BERT on our target FiQA task corpus has slight improvements in the MRR and NDCG. As pointed out in \cite{araci}, further pre-training BERT on a large financial domain corpus did not improve the sentiment classification task significantly mostly likely because the performance of $\text{BERT}_{\textsc{BASE}}$ was already good that there was not much room for improvement. \cite{araci} sugested that this may only be the case for short sentence classification and that more experiments with labeled financial dataset is necessary to make a conclusion about the effects of  further pre-training. Our experiments support \cite{araci}'s idea that classifiers improve only slightly with further pre-training and show that this is the case even for long sequences.

\section{Further Fine-tuning (RQ3)}

The results in Table \ref{tab:result} show that further fine-tuning BERT using the Transfer and Adapt approach have the highest MRR and NDCG. This suggests that for the financial QA task, further fine-tuning is more effective than further pre-training BERT on a domain or task-specific corpus. A possible explanation for the effectiveness of further fine-tuning is that the subset of the MS MACRO corpus (12.8 million QA pairs) used in the transfer fine-tuning step is about 45 times bigger than the target FiQA dataset (283,707 QA pairs). Moreover, the MS MACRO passage retrieval corpus already contained questions corresponding to the financial domain, such as "what is a bank transit number" and "how much can i contribute to nondeductible ira". Hence, the adaptation fine-tuning step has an advantage because more domain-specific knowledge is provided from another data source.

\section{Baseline and Advanced Models Comparison (RQ4)}

The results of our implementation of QA-LSTM scored the lowest on all metrics. However, it should be noted that we focus our experiments on fine-tuning BERT, therefore the QA-LSTM model was not thoroughly experimented with. Nevertheless, \cite{tran} implemented two Siamese-based LSTM models for the FiQA dataset, namely SRanker$_{mlp}$ and SRanker$_{bilinear}$, which compute the attention mechanisms using multi-layer perceptrons and a bilinear term respectively. \cite{tran}'s Siamese-based LSTM models and the SOTA CARanker model based on the Compare-Aggregate architecture provide us the performance of using representation learning for the financial QA task.

Interestingly, the results in Table \ref{tab:result} show that Anserini's BM25 implementation already outperforms \cite{tran}'s  CARanker by 2.6\% on MRR, 5.3\% on NDCG, and 7.1\% on Precision@1. This baseline IR method, therefore, is more effective and powerful compared to complicated deep representation learning methods in our experiments. Moreover, the results show that representation learning is not suitable for context-specific corpora.

The combination of BM25 and fine-tuned BERT models yield the best results. FinBERT-Domain, a model further pre-trained BERT on a large financial corpus, achieved the best Precision@1 that is 22\% better than the SOTA. Moreover, FinBERT-QA, a BERT model using the Transfer and Adapt further fine-tuning approach achieved the best MRR@10 and NDCG@10 results, where the MRR is 16\% and the NDCG is 17\% better than the SOTA. Overall, using pre-trained BERT models improved the SOTA results of the FiQA dataset by an average of 18\% on MRR, NDCG, and Precision. This improvement can be seen in Table \ref{tab:QA}, where the ground truth answer was returned at the first position by FinBERT-QA and the negative answer was returned at the first position by the Answer Retriever.

\section{Limitations}

Even though our fine-tuned BERT models achieved good results, we address the following limitations: 

\begin{itemize}
\item \textbf{Answer Retriever:} It should be noted that there are missing relevant answers in top 50 candidate answers returned by the Answer Retriever. In particular, 19\% of the questions in the test set didn't have \textit{any} relevant answers and 50\% of the questions didn't have \textit{all} the relevant answers. This limitation could have greatly effected the performance of our Answer Re-ranker and motivates more research on advanced approaches to improve the answer retrieval step.
\item \textbf{Maximum Sequence Length Restriction:} Since the pre-trained BERT model used a maximum input sequence length of 512, the concatenation of question, answer, and special tokens are truncated to have a maximum of 512 tokens. This becomes a limitation for tasks with long input sequences. In the FiQA dataset, 2\% of the answers have more than 512 tokens. Therefore, the missing content of some sequences could have affected the performance. The effects of the maximum sequence length for our fine-tuned BERT models on MRR, NDCG, and Precision are shown in Figure \ref{fig:max}. The evaluation metrics are affected by the maximum sequence length by a significant amount for all models. Hence, further research into solving this limitation is essential  for tasks that have long input sequences.

\begin{figure}[h!]
\centering
\includegraphics[width=.5\textwidth]{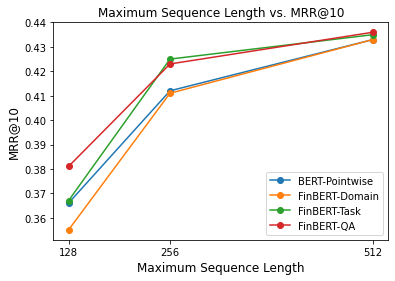}\hfill
\includegraphics[width=.5\textwidth]{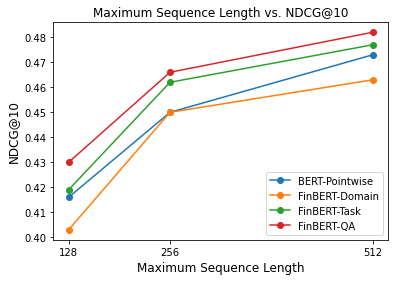}\\
\includegraphics[width=.5\textwidth]{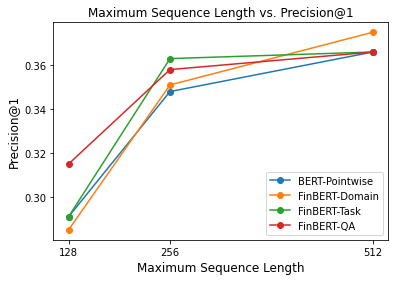}
\caption{Comparison of the effects of the maximum sequence length on the evaluation metrics of the fine-tuned BERT models. The evaluation metrics are affected by the maximum sequence length by a significant amount for all models.}
\label{fig:max}
\end{figure} 
\item \textbf{Data Quality:} A detrimental limitation of our models is the data quality of the FiQA dataset. Since the dataset was obtained by crawling community QA websites, a subset of the answers are based on opinions. Hence, the answers might not necessarily answer the question. For example, Table \ref{tab:QA_bad} shows a poorly labeled ground truth answer that does not answer the question. Moreover, the same or similar questions could be crawled from different QA websites, therefore, the answers could be correct even if they are not in the set of relevant answers due to different wording of the questions. For example, Table \ref{tab:QA_bad2} shows an answer returned by FinBERT-QA that was classified as incorrect even though it answers the question. Hence, further work on obtaining high quality labeled financial QA datasets is critical.

\end{itemize}

\begin{center}
\begin{minipage}{\linewidth}
\centering
\renewcommand{\arraystretch}{1.5}
\begin{tabular}{||>{\arraybackslash}p{12cm}||}

            \hline
            \textbf{Question:}  What is a good open source Windows finance software?\\ \hline
             \textbf{Ground truth answer:} Have you tried others on Wikipedia's list?\\ \hline              
\end{tabular}
\captionof{table} {Example of poorly labeled ground truth answer in the FiQA dataset}
\label{tab:QA_bad}
\end{minipage}
\end{center}

\begin{center}
\begin{minipage}{\linewidth}
\centering
\renewcommand{\arraystretch}{1.5}
\begin{tabular}{||>{\arraybackslash}p{12cm}||}

            \hline
            \textbf{Question:}  Where do large corporations store their massive amounts of cash?\\ \hline
             \textbf{Ground truth answer:} Short term investments, treasuries, current accounts\\ \hline
             \textbf{FinBERT-QA answer:} For individuals, it's very easy to just put your cash in a checking account.  Checking accounts are insured and non-volatile.  But what happens when you're GE or Apple or Panama?   You can't just flop a couple billion dollars in to a Chase checking account and call it a day.  Although, you still need a safe place to store money that won't be terribly volatile.  GE can buy a billion dollars of treasury bonds.  Many companies need tremendous amounts of collateral on hand, amounts far in excess of the capacity of a checking account; those funds are stored in treasuries of some sort. \\ \hline            
\end{tabular}
\captionof{table} {Example of an answer returned by FinBERT-QA that was classified as incorrect even though it answers the question.}
\label{tab:QA_bad2}
\end{minipage}
\end{center}

\chapter{Conclusion}\label{chap:conclusion}

In this work, we apply fine-tuned BERT models to non-factoid QA in the financial domain for the first time to the best of our knowledge. Using pre-trained BERT models helped mitigate  the drawbacks of data scarcity, financial language specificity, and shallow pre-trained static word embeddings usage of traditional deep learning methods.

We formulate the non-factoid answer selection problem as a re-ranking task and build a financial QA system by first using Anserini's BM25 implementation as the Answer Retriever to obtain a list of candidate answers for each question. Then we experiment with different models for the Answer Re-ranker to return the most relevant questions. FinBERT-QA, built from first transferring and fine-tuning a pre-trained BERT model to a large QA corpus then adapting the transferred model to the target FiQA dataset achieved the state-of-the-art results for task 2 of the FiQA challenge and improved the benchmarks significantly. 

Our research deepens the general understanding of fine-tuning pre-trained BERT models for financial non-factoid QA, including the effectiveness of various learning, pre-training, and fine-tuning approaches.

\section{Future Work}

The limitations from the experimental results motivate future work in investing more research on advanced methods to improve the answer retrieval step, solving the maximum sequence length restriction of BERT, and curating a higher quality labeled financial QA dataset. Another interesting future work idea is to conduct more experiments with other financial and opinion-based datasets. The large financial corpus that our experiments of further pre-training BERT was based on was derived from financial news stories. Since the target FiQA dataset is based on opinions, further pre-training BERT on a large opinion-based dataset could be done to capture a less formal form of language.

Pre-trained Transformer-based language models have been an active research in the NLP community due to their effectiveness. More recent pre-trained language models such as XLNet \cite{xlnet} and RoBERTa \cite{roberta} were shown to have outperformed BERT. In particular, XLNet is a new architecture that is pre-trained using approximately 10 times more data than BERT and using a batch size that is eight times larger. Motivated by XLNet, RoBERTa, which stands for Robustly optimized BERT approach, also retrained BERT for longer and with bigger batches over more data. In particular, RoBERTa uses 160 GB of text for pre-training including the 16 GB of the original data used to pre-train BERT. As a result, RoBERTa outperforms both BERT and XLNet,  achieving the state-of-the-art results on multiple NLP tasks. Particuarly, \cite{tanda} achieved the current state-of-the-art results on the TREC-QA and WikiQA datasets by fine-tuning the $\text{RoBERTa}_{\textsc{LARGE}}$ model using the Transfer and Adapt method. Hence, it would be interesting to first transfer and fine-tune a pre-trained RoBERTa model to a large QA dataset like MS MACRO, then adapt the transfered model to the FiQA dataset to evaluate the results.

\nocite{*}

\printbibliography

\end{document}